\pdfoutput=1

\documentclass[11pt]{article}

\usepackage[preprint]{acl}

\usepackage{times}
\usepackage{latexsym}

\usepackage[T1]{fontenc}

\usepackage[utf8]{inputenc}

\usepackage{microtype}

\usepackage{inconsolata}

\usepackage{graphicx}

\usepackage{amsmath}
\usepackage{amsfonts}
\usepackage{bm}
\usepackage{subcaption}
\usepackage{booktabs}
\usepackage{multirow}
\usepackage{xspace}
\usepackage{comment}
\usepackage{scalerel}

\usepackage{arydshln}
\makeatletter
\def\adl@drawiv#1#2#3{%
        \hskip.5\tabcolsep
        \xleaders#3{#2.5\@tempdimb #1{1}#2.5\@tempdimb}%
                #2\z@ plus1fil minus1fil\relax
        \hskip.5\tabcolsep}
\newcommand{\cdashlinelr}[1]{%
  \noalign{\vskip 2pt
           \global\let\@dashdrawstore\adl@draw
           \global\let\adl@draw\adl@drawiv}
  \cdashline{#1}[.4pt/2pt]
  \noalign{\global\let\adl@draw\@dashdrawstore
           \vskip 2pt}}
\makeatother

\DeclareMathOperator*{\oprod}{\scalerel*{\odot}{\prod}}

\newcommand{\methodname}{\textsc{LaTIM}\xspace}

\title{\methodname: \\ Measuring Latent Token-to-Token Interactions in Mamba Models}

\author{
 \textbf{Hugo Pitorro},
 \textbf{Marcos Treviso}
\\
 Instituto de Telecomunicações, Lisbon
\\
 \small{
    \href{mailto:hugo.pitorro@gmail.com}{\texttt{hugo.pitorro@gmail.com}}
 }
}

\begin{document}
\maketitle
\begin{abstract}

State space models (SSMs), such as Mamba, have emerged as an efficient alternative to transformers for long-context sequence modeling. 
However, despite their growing adoption, SSMs lack the interpretability tools that have been crucial for understanding and improving attention-based architectures. 
While recent efforts provide insights into Mamba's internal mechanisms, they do not explicitly decompose token-wise contributions, leaving gaps in understanding how Mamba \emph{selectively} processes sequences across layers. 
In this work, we introduce \methodname, a novel token-level decomposition method for both Mamba-1 and Mamba-2 that enables fine-grained interpretability. 
We extensively evaluate our method across diverse tasks, including machine translation, copying, and retrieval-based generation, demonstrating its effectiveness in revealing Mamba's token-to-token interaction patterns. Our code is available at \url{https://github.com/deep-spin/latim}.

\end{abstract}

\section{Introduction}

State space models (SSMs), such as S4~\citep{gu2022efficiently}, have emerged as a promising alternative to transformers for long-context modeling. 
Unlike transformers~\citep{vaswani2017attention}, which explicitly compute pairwise token interactions and require quadratic memory, SSMs leverage structured recurrence mechanisms that enable more efficient sequence processing. 
Among them, the Mamba architecture \citep{gu2023mamba, dao2024transformersssmsgeneralizedmodels} has demonstrated strong performance in language modeling and other modalities while significantly reducing runtime and memory requirements~\citep{xu2024survey}. 
Additionally, hybrid architectures that integrate both Mamba and attention mechanisms often outperform purely homogeneous models by combining the efficiency of recurrence with the expressivity of attention~\citep{lenz2025jamba, dong2025hymba, pitorro-etal-2024-effective}. While these findings highlight the relevance of Mamba models, their internal decision-making processes remain opaque, hindering their reliability.

Interpretability techniques have played a key role in the widespread adoption of transformers, enabling researchers to analyze token interactions and information flow~\citep{mohebbi-etal-2024-transformer,ferrando-etal-2024-primer}.
However, in contrast to transformers, where attention scores offer a direct visualization of how the model distributes importance across tokens, Mamba lacks an explicit mechanism to reveal where it is ``attending'' at each step. 
Existing interpretability efforts for Mamba attempt to bridge this gap by reformulating its computations into attention-like representations. 
For instance, \text{MambaAttention}~\citep{ali2024hiddenattentionmambamodels} reformulates the model's computation in terms of implicit attention matrices, 
while \text{MambaLRP}~\citep{jafari2024mambalrp} uses layer-wise propagation analysis to track gradient flow.
However, these methods do not explicitly decompose contributions into fine-grained elements across layers, leaving gaps in understanding how Mamba \emph{selectively} processes sequences.

In this work, we bridge this gap by introducing \methodname, a novel token-level decomposition method for both Mamba-1 and Mamba-2. 
Our approach reformulates the SSM computation to enable token-by-token analysis, allowing us to adapt attention-based interpretability techniques, such as ALTI~\citep{ferrando-etal-2022-measuring}, to the Mamba architecture.
We extensively evaluate our method across diverse tasks, including the copying task \citep{jelassi2024repeat} in \S\ref{sec:copying}, which features a well-defined diagonal attention pattern; 
machine translation in \S\ref{sec:mt}, where precise source$\leftrightarrow$target alignment is essential; 
and retrieval-based generation \citep{hsieh2024rulerwhatsrealcontext} in \S\ref{sec:retrieval_based_generation}, where ground-truth context allows direct evaluation of token importance.
Our method not only improves Mamba's interpretability but also defines a robust framework for analyzing token interactions in SSMs, paving the way for more transparent models.\looseness=-1

\section{Background}

\subsection{Transformers} 
\label{subsec:transformers}

A key component in the transformer architecture is the attention mechanism, which is responsible for mixing input sequences $\bm{X} = \langle \bm{x}_1, ..., \bm{x}_N \rangle$, where each $\bm{x}_i \in \mathbb{R}^{D}$. 
Concretely, given query
$\bm{Q}^h = \bm{X} \bm{W}_q^h \in \mathbb{R}^{N \times D'}$,  key $\bm{K}^h = \bm{X} \bm{W}_k^h \in \mathbb{R}^{N \times D'}$, and  value $\bm{V}^h = \bm{X} \bm{W}_v^h \in \mathbb{R}^{N \times D'}$ matrices as input, where $1 \leq h \leq H$ is the head dimension,
the \emph{multi-head attention mechanism} is defined as follows~\citep{vaswani2017attention}:
\begin{equation}\label{eq:attention}
    \textsf{Attn}(\bm{X})_h = 
    \underbrace{\pi
    \Bigg(
            \frac{\bm{Q}^h\bm{K}^{h^\top}}{\sqrt{D'}}
    \Bigg)}_{\bm{A}^h \in \mathbb{R}^{N \times N}} \bm{V}^h
    \in \mathbb{R}^{N \times D'},
\end{equation}
where $\pi$ maps rows to distributions, with $\pi := \textsf{softmax}$ being a common choice.

\paragraph{Transformer block.} The attention is combined with other modules in order to form a transformer block. The full block, with pre LayerNorm~(\textsf{LN}, \citealt{ba2016layer}), can be described as follows:
\begin{align}
    \bm{X}_l &= \textsf{LN}(\bm{X}) &\in \mathbb{R}^{N \times D}, \\
    \bm{Y}_a &= \textsf{Concat}(\textsf{Attn}(\bm{X}_l)_h \bm{W}^h_o), &\in \mathbb{R}^{N \times D},\nonumber \\
    \bm{Y} &= \bm{Y}_a + \bm{X} &\in \mathbb{R}^{N \times D},\nonumber
\end{align}
where we denote $\textsf{Concat}(\cdot)$ as the concatenation of all heads $1 \leq h \leq H$, and $\bm{W}_o \in \mathbb{R}^{D' \times D}$.
In words, the attention output is projected through $\bm{W}^h_O$ and, together with a residual stream and pre-layer norm, forms the output of the block.

\subsection{Attention Decomposition}
\label{subsec:attention_decomposition}

Transformers benefit from attention maps for interpretability, but these do not fully capture token influence on predictions. 
Token attribution methods address this by decomposing the forward pass into token-wise contributions~\citep{kobayashi-etal-2021-incorporating}. This section presents two key approaches---direct token-to-token decomposition and logit attribution---which motivate our interpretability method for Mamba.

\paragraph{Token Contributions.} 
To determine the influence of token $j$ on the representation of token $i$, we express the output at a certain layer as follows:\footnote{We ignore the bias terms for clarity (w.l.o.g). Moreover, in a decoder-only model we have $1 \leq j \leq i$.}
\begin{equation} \label{eq:decomposed_yi_transformer}
\bm{y}_i = \sum_{j=1}^N T_i(\bm{x}_j) \in \mathbb{R}^{D},
\end{equation}
where the transformed contribution of $\bm{x}_j$ to $\bm{y}_i$ is
\begin{equation} \label{eq:transformer-tij}
T_i(\bm{x}_j) = \sum_{h=1}^H \bm{W}_o^h \bm{A}_{i,j}^h \bm{W}_v^h \cdot \textsf{LN}(\bm{x}_j) + \delta_{i,j}\bm{x}_i,
\end{equation}
with $\delta_{i,j}$ denoting the Kronecker delta.

\paragraph{Token-to-Token Importance.} 
Using this decomposition, we can obtain token-to-token importance scores via vector norms~\citep{kobayashi-etal-2021-incorporating}:
\begin{equation}
    C_{i,j} = \left\|T_i(\bm{x}_j) \right\|_2,
\end{equation}
or via ALTI's contextual mixing approach~\citep{ferrando-etal-2022-measuring}: 
\begin{align} \label{eq:transformer-alti}
    C_{i,j} &= \frac{\left[\|\bm{y}_i\|_1 - \|\bm{y}_i - T_i(\bm{x}_j) \|_1 \right]_+}{\sum_k \left[\|\bm{y}_i\|_1 -\|\bm{y}_i - T_i(\bm{x}_k) \|_1 \right]_+},
\end{align}
where $[\cdot]_+$ represents the ReLU function.

\paragraph{Logit Contributions.}

While token-wise decomposition methods capture interactions within a layer, they do not measure a token's direct contribution to the final output. To bridge this gap, ALTI-Logit~\citep{ferrando-etal-2023-explaining} traces token contributions through the residual stream up to the final prediction.
Formally, given a token $w(i) \in \mathcal{V}$, the contribution of token $j$ at layer $l$ is given by:
\begin{equation} 
\Delta_{i,j}^{(l)} = T_{i}^{(l)}(\bm{x}_{j}^{(l-1)})^\top \bm{U}_{w(i)}, 
\end{equation}
where $\bm{U} \in \mathbb{R}^{|\mathcal{V}| \times D}$ is the output embedding matrix. 
Let $\bm{R}^{(l)} = \bm{P}^{(l)}\cdots \bm{P}^{(2)} \bm{P}^{(1)}$ denote the residual stream at layer $l$, where $P^{(l)}_{i,j}$ refers to the contribution of $\bm{x}_i^{(l-1)}$ to $\bm{x}_j^{(l)}$ such that $\sum_j P^{(l)}_{i,j} = 1$.
Then, the final pairwise contribution score aggregated from all $L$ layers is
\begin{equation} \label{eq:transformer-alti-logit}
     C_{i,j} = \sum_{l=1}^L \Delta_{i,j}^{(l)} \bm{R}_j^{(l-1)}. 
\end{equation}
ALTI-Logit provides a final-layer attribution score, making it particularly useful for output-sensitive interpretability. 
In Section~\ref{sec:method}, we follow these principles to design attribution methods for Mamba.

\subsection{State Space Models (SSMs)}

SSMs \citep{gu2020hippo} are a type of sequence mixing layer that  
process sequences through a linear recurrence.
Letting $\bm{H}_i \in \mathbb{R}^{R \times D}$ denote the ``state'' at the $i\textsuperscript{th}$ time step, a discrete SSM can be formulated as follows~\citep{pitorro-etal-2024-effective}:\footnote{A discretization step is required to obtain discrete parameters (e.g., via the zero-order hold rule); 
however, we follow \citet{pitorro-etal-2024-effective} and omit this step for clarity.}
\begin{align}
  \label{eq:discrete-ssm}
    \bm{H}_i &= \bm{A}\bm{H}_{i-1} + \bm{b} \bm{x}_i^\top &\in \mathbb{R}^{R \times D}, \\
    \bm{\upsilon}_i &= \bm{H}_i^\top \bm{c} + \bm{D}\bm{x}_i &\in \mathbb{R}^{D}, \nonumber
\end{align}
where $\bm{A} \in \mathbb{R}^{R \times R}$, $\bm{b} \in \mathbb{R}^{R}$, $\bm{c} \in \mathbb{R}^{R}$, $\bm{D} \in \mathbb{R}^{D \times D}$ are (discrete) parameters shared for all $i$.%

\paragraph{Mamba-1.}

The first version of Mamba~\citep{gu2023mamba} extends the previous formulation into an \emph{input-dependent} SSM by turning the parameters into learnable projections of the current input $\bm{x}_i$:
\begin{align}
  \label{eq:discrete-mamba1}
    \bm{H}_i &= \bm{A}_i \odot \bm{H}_{i-1} + \bm{B}_i \odot \bm{X}_i &\in \mathbb{R}^{R \times D}, \\
    \bm{\upsilon}_i &= \bm{H}_i^\top \bm{c}_i + \bm{D}\bm{x}_i &\in \mathbb{R}^{D}, \nonumber 
\end{align}
where $\bm{X}_i = \bm{1}_r\bm{x}_i^\top \in \mathbb{R}^{R \times D}$ is an $R$-sized stack of the input, $\bm{A}_i \in \mathbb{R}^{R \times D}$ represents $D$ diagonal matrices of size $R \times R$, $\bm{B}_i \in \mathbb{R}^{R \times D}$, $\bm{c}_i \in \mathbb{R}^{R}$, and $\odot$ is the Hadamard product.

\paragraph{Mamba-1 block.} 
Analogously to transformers, the Mamba-1 model is a collection of stacked blocks containing a sequence mixing layer and a gating mechanism. 
Concretely, the sequence mixing layer can be fully described as:
\begin{align} \label{eq:mamba1_forward_block}
  \bm{\Psi} &= \textsf{Conv1D}(\bm{X} \bm{W}_x) &\in \mathbb{R}^{N \times 2D}, \\
  \bm{\Phi} &= \textsf{SiLU}(\bm{\Psi}) &\in \mathbb{R}^{N \times 2D}, \nonumber \\
  \bm{A},\  &\bm{B}, \bm{C} = \textsf{Linear}(\bm{\Phi}) &\in \mathbb{R}^{N \times R}, \nonumber \\
  \bm{\Upsilon} &= \textsf{SSM} (\bm{\Phi}; \bm{A}, \bm{B}, \bm{C}, \bm{D}) &\in \mathbb{R}^{N \times 2D}, \nonumber
\end{align}
where $\bm{W}_x \in \mathbb{R}^{D \times 2D}$ and $\textsf{Linear}: \mathbb{R}^{N \times 2D} \to \mathbb{R}^{N \times R}$ represents a set of low-rank projections.
The gating mechanism is employed as follows:
\begin{align} \label{eq:mamba1_gating}
  \bm{Z} &= \textsf{SiLU}(\bm{X} \bm{W}_z) &\in \mathbb{R}^{N \times 2D}, \\
  \bm{U} &= \bm{\Upsilon} \odot \bm{Z} &\in \mathbb{R}^{N \times 2D}, \nonumber \\
  \bm{Y} &= \bm{U} \bm{W}_o &\in \mathbb{R}^{N \times D}, \nonumber
\end{align}
where $\bm{W}_z \in \mathbb{R}^{D \times 2D}$ and  $\bm{W}_o \in \mathbb{R}^{2D \times D}$.

\paragraph{Mamba-2.}
Mamba-2~\citep{dao2024transformersssmsgeneralizedmodels} introduces a simpler SSM formulation by defining $\bm{A}$ as a scalar times identity $\bm{A}_i = a_i\bm{I}_{R \times R}$.  This leads to the following \emph{input-dependent} model:
\begin{align}
    \label{eq:discrete-mamba2}
    \bm{H}_i &=\bm{A}_i \bm{H}_{i-1} + \bm{B}_i \odot \bm{X}_i &\in \mathbb{R}^{R \times D}, \\
    \bm{\upsilon}_i &= \bm{H}_i^\top \bm{c}_i + \bm{D}\bm{x}_i &\in \mathbb{R}^{D}. \nonumber 
\end{align}
In contrast to Mamba-1 (\textit{c.f.} Equation~\ref{eq:discrete-mamba1}), the input-dependent parameter $\bm{A}_i \in \mathbb{R}^{R \times R}$ represents a single diagonal matrix.

\paragraph{Mamba-2 block.}
Regarding block structure, Mamba-2 draws the parameters $\bm{A}, \bm{B}, \bm{C}$ directly from the initial input $\bm{X}$, and further introduces a GroupNorm layer~\citep{Wu_2018_ECCV} after the gating mechanism for additional stability:
\begin{align} \label{eq:mamba2_forward_block}
  \bm{U} &= \textsf{GroupNorm}(\bm{\Upsilon} \odot \bm{Z}) &\in \mathbb{R}^{N \times 2D}. 
\end{align}

\subsection{Hidden Attention in Mamba}

As noted by \citet{ali2024hiddenattentionmambamodels} and \citet{dao2024transformersssmsgeneralizedmodels}, by unrolling Mamba's recurrence we can interpret the sequence mixing layer as multiplying a lower-triangular matrix $\bm{M}$ with the entire input $\bm{\Upsilon} = \bm{M}\bm{X}$ (independently for each channel/head). 
More generally, by unrolling Mamba-1's recurrence defined in Equations~\ref{eq:discrete-mamba1}, we can show that $\bm{M}_{i,j} \in \mathbb{R}^{D \times D}$ has the following form:
\begin{equation}
\bm{M}_{i,j} = 
    \textsf{Diag} \left( \left[ \left(\oprod_{k=j+1}^{i} \bm{A}_k \right) \odot \bm{B}_j \right]^{\top} \bm{c}_i \right),
\label{eq:mamba-attention}
\end{equation}
for all $j \leq i$, and $\bm{M}_{i,j} = \bm{0}$ otherwise. 
A similar expression can be derived for Mamba-2 by noticing that $\bm{A}_k \in \mathbb{R}^{R \times R}$ is, by definition, a diagonal matrix. 
Importantly, for each dimension $d \in [D]$, this is an \emph{implicit attention} matrix akin to transformers' attention matrix.
We provide more details on this derivation in \S\ref{sec:derivation_m}.

\section{\methodname}
\label{sec:method}

While the attention mechanism found in transformers allows us to decompose the contributions of different input tokens, decomposing individual token contributions is challenging for Mamba.
Additionally, in Mamba-1 the channel dimensionality is often large in practice, and therefore manual inspection of all attention maps per layer and sample quickly becomes unfeasible (e.g., a 370M model has 48 layers with $D=1024$).
Although Mamba-2 alleviates this issue by using a smaller number of heads, it remains unclear how to obtain a single attention plot for each layer or for an entire sample.
Overall, our goal is to rearrange the forward pass from both Mamba-1 and Mamba-2 such that we can measure the total contribution of token $\bm{x}_j$ towards the output $\bm{y}_i$, similar to the definition of $T_i(\bm{x}_j)$ in Equation~\ref{eq:transformer-tij} tailored for transformers.

\subsection{Mamba-1 Decomposition}
\label{subsec:method_mamba1}

In this direction, we start by revisiting Mamba's forward pass at step $i$ in Equation~\ref{eq:mamba1_forward_block}. 
The first component of Mamba-1 block is the 1D convolution layer. Concretely, letting $w \in \mathbb{N}$ denote the kernel size, the 1D causal convolution output for a token $i$ can be described as:
\begin{align}
    \bm{\psi}_i &= \mathrm{Conv1D}\left(\bm{X}\bm{W}_x; w \right)_i \\
    &= \sum_{k=1}^w  \bm{W}_{c}^{(k)} \left( \bm{W}_x^\top \bm{x}_{i-w+k} \right) + \bm{b}_c,
\end{align}
where $\bm{W}^{(k)}_{c} \in \mathbb{R}^{d \times d}$ and $\bm{b}_c$ represents the convolution kernel and bias, respectively. 
Next, $\bm{\psi}_i$ is transformed via a \textsf{SiLU} activation $\bm{\phi}_i = \textsf{SiLU}(\bm{\psi}_i)$, which, in turn, is passed to the SSM module, $\bm{\upsilon}_{i} = \textsf{SSM}(\bm{\phi}_i)$. 
Therefore, in order to compute the contribution of token-to-token interactions, we first need to unroll the SSM recurrence from Equation~\ref{eq:discrete-mamba1}. 
To that end, we leverage the tensor $\bm{M}$ defined in Equation~\ref{eq:mamba-attention} and treat the term $\bm{D}\bm{\phi}_i$ as a skip-connection, leading to:
\begin{equation} \label{eq:silu_placement}
    \bm{\upsilon}_i = \sum_{j = 1}^i (\bm{M}_{i,j} + \delta_{i,j}\bm{D}) \underbrace{\bm{\phi}_j}_{ \textsf{SiLU}\left( \bm{\psi}_j \right)},
\end{equation}
where $\delta_{i,j}$ is the Kronecker delta. 
Unfortunately, the non-additivity of the $\textsf{SiLU}$ activation prevents the decomposition of $\bm{\upsilon}_i$ as a sum of previous token interactions. 
That is, we cannot rearrange the above expression such that we use the $j$\textsuperscript{th} token only at the $j$\textsuperscript{th} iteration, prohibiting us from deriving token-to-token contributions as done in transformers (see Section~\ref{subsec:attention_decomposition}).
However, if we assume the existence of an additive function $f$ that approximates well the \textsf{SiLU} activation, we can decompose $\bm{\phi}_j$ as follows:\footnote{We explicitly include $\delta_{j,0}$ into the expression to account for the convolution bias, which is only added once per channel.}
\begin{equation}
\label{eq:f_additive_decomp_mamba}
    \bm{\phi}_j = \sum_{k=1}^w f ( \underbrace{ \bm{W}_c^{(k)} \bm{W}_x^\top \bm{x}_{j-w+k} + \delta_{k,0} \bm{b}_c }_{\bm{\varphi}_j^{(k)}} ). 
\end{equation}
This decomposition allows us to derive a more \emph{interpretable} output for Mamba's recurrent module:
\begin{align} \label{eq:tijd_mamba1}
    \bm{\upsilon}_i = \sum_{j=1}^{i} \sum_{k=1}^w \left( \bm{M}_{i,j+k} + \delta_{i,j+k} \bm{D} \right) \bm{\varphi}_j^{(k)}.
\end{align}
Importantly, we can modify the above expression in order to obtain the vector representation that stems from interactions with the $j$\textsuperscript{th} token as follows:
\begin{align} \label{eq:tijd_mamba1_decomp}
    \bm{\upsilon}_{i \leftarrow j} =  \sum_{k=1}^w \left( \bm{M}_{i,j+k} + \delta_{i,j+k} \bm{D} \right) \bm{\varphi}_{j}^{(k)}.
\end{align}
Finally, after considering the gating mechanism and the output projection from Equation~\ref{eq:mamba1_gating}, we obtain the $(i, j)$ contribution vector:
\begin{equation}
    T_i(\bm{x}_j) =  
    \bm{W}_o^{\top} \left( \bm{Z}_i \odot \bm{\upsilon}_{i \leftarrow j} \right). 
\end{equation}
And similarly to the way attention is decomposed in transformers (see Equations~\ref{eq:decomposed_yi_transformer} and \ref{eq:transformer-tij}), the final output can be computed by integrating the contribution from all previous tokens:
\begin{equation}
    \bm{y}_i = \sum_{j=1}^i T_i(\bm{x}_j).
\end{equation}

\begin{table}[t]
    \centering
    \small
    \setlength{\tabcolsep}{3pt} %
    \begin{tabular}{ll}
    \toprule
    \bf Method & \bf Expression \\
    \midrule
    \methodname ($\ell_p$)    & $C_{ij} = \|T_i(\bm{x}_j)\|_p$ \\
    \methodname (ALTI)        & $C_{ij} \propto \left[\|\bm{y}_i\|_1 - \|\bm{y}_i - T_i(\bm{x}_j) \|_1 \right]_+$ \\
    \methodname (ALTI-Logit)  & $C_{ij} = T_{i}(\bm{x}_{j})^\top \bm{U}_{w(i)} \bm{R}_j$ \\
    \bottomrule
    \end{tabular}
    \caption{Overview of \methodname-based methods for obtaining contribution scores for $(i,j)$ token interactions. }
    \label{tab:equations_overview}
\end{table}

\subsection{Mamba-2 Decomposition}
\label{subsec:method_mamba2}

Recall from Equation~\ref{eq:mamba2_forward_block} that Mamba-2 places a \textsf{GroupNorm} layer on the output of the SSM module.  
Let $\bm{\upsilon}_i \in\mathbb{R}^{2D}$ be the SSM output at token $i$, 
and define $\bm{u}_i = \bm{Z}_i \odot \bm{\upsilon}_i$.  
At test time, \textsf{GroupNorm} can be viewed as an affine map around $\bm{u}_i$, 
\begin{equation}
  \textsf{GroupNorm}(\bm{u}_i ) = \gamma_i(\bm{u}_i)\,\bm{u}_i + \beta_i,
\end{equation}
where $\gamma_i(\bm{u}_i)$ is a (fixed) linear operator once $\bm{u}_i$ is known, and $\beta_i$ is an offset.\footnote{We follow \citet{ferrando-etal-2022-measuring} and ignore the offset term as it not attributed to any token.}  
Hence, if $\bm{u}_{i\leftarrow j}$ denotes the portion of $\bm{u}_i$ that originates from token~$j$, 
its contribution passes through GroupNorm in the same linear fashion.
Finally, applying the output projection $\bm{W}_o$ yields the token decomposition:
\begin{equation}
  T_i(\bm{x}_j)
  = \bm{W}_o^\top
  \Bigl[\gamma_i\bigl(\bm{u}_i\bigr)\,\bm{u}_{i\leftarrow j} \Bigr].
\end{equation}
As we are interested in obtaining token-to-token interpretability scores, we can apply various scalar aggregation functions to $T_i(\bm{x}_j)$.  Common examples include $\ell_1$ or $\ell_2$ norms~\citep{kobayashi-etal-2021-incorporating}, as well as the ALTI~\citep{ferrando-etal-2022-measuring} and ALTI-Logit~\citep{ferrando-etal-2023-explaining} approaches. 
We provide a summary of \methodname variants that leverage these aggregations in Table~\ref{tab:equations_overview}.

\subsection{Decomposition Error}
\label{subsec:decomposition_error}

\paragraph{Approximated Strategy.}  
Unlike attention decomposition in transformers, Mamba requires an additive function $f$ in Equation~\ref{eq:f_additive_decomp_mamba} to linearly decompose pairwise interactions. 
Ideally, $f$ should closely approximate the original non-additive expression $\bm{\phi}_i = \textsf{SiLU}(\bm{\psi}_i)$. 
To assess this, we explore different approximation strategies in \S\ref{sec:extended-error}, including first- and second-order Taylor expansions around zero. 
Surprisingly, we find that directly setting $f$ as \textsf{SiLU} yields the lowest approximation error across all layers. 
Therefore, unless explicitly stated otherwise, \methodname refers to our decomposition method using $f := \textsf{SiLU}$.

\paragraph{Exact Strategy.}  
While a well-chosen approximation function $f$ enables interpretability without requiring model retraining, it does not fully recover the exact Mamba block's output. 
To eliminate this discrepancy, we propose a modified version of Mamba that removes the \textsf{SiLU} activation, simplifying the computation to $\bm{\phi}_i = \bm{\psi}_i$, which effectively turns $f$ into the identity function in Equation~\ref{eq:f_additive_decomp_mamba}. 
Though this approach requires retraining, we demonstrate in Section~\ref{subsec:approximation_error_analysis} that it achieves zero decomposition error while maintaining the same level of interpretability and task performance.

\begin{figure}[t]
  \centering
  \begin{subfigure}{0.23\textwidth}
    \includegraphics[width=\textwidth]{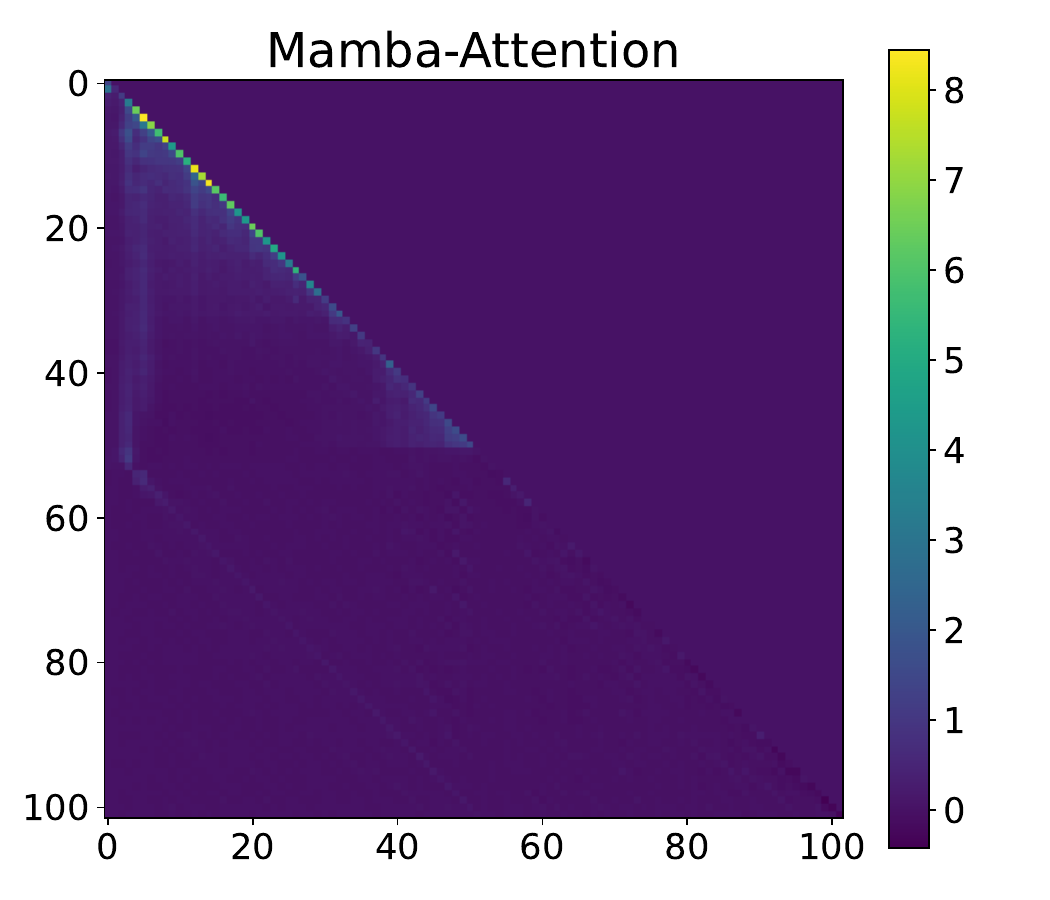}
  \end{subfigure}
  \hfill
  \begin{subfigure}{0.23\textwidth}
    \includegraphics[width=\textwidth]{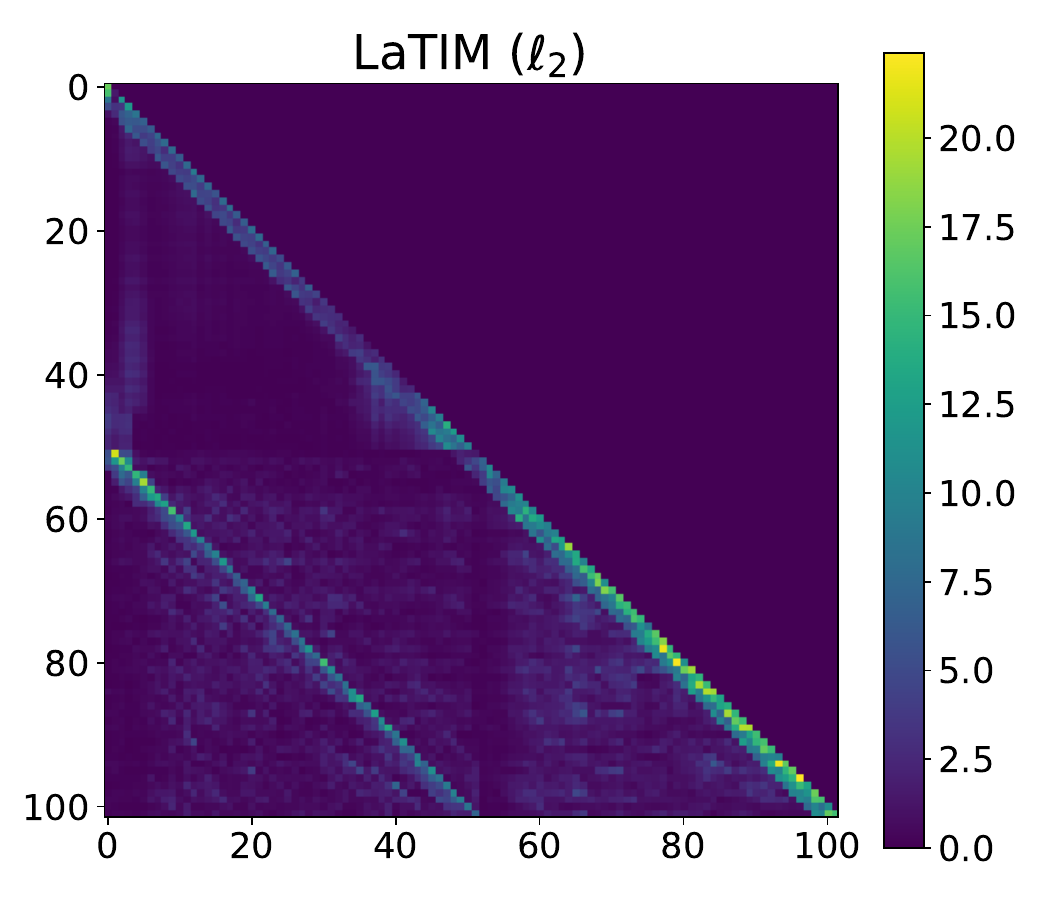}
  \end{subfigure}
  \hfill
  \begin{subfigure}{0.23\textwidth}
    \includegraphics[width=\textwidth]{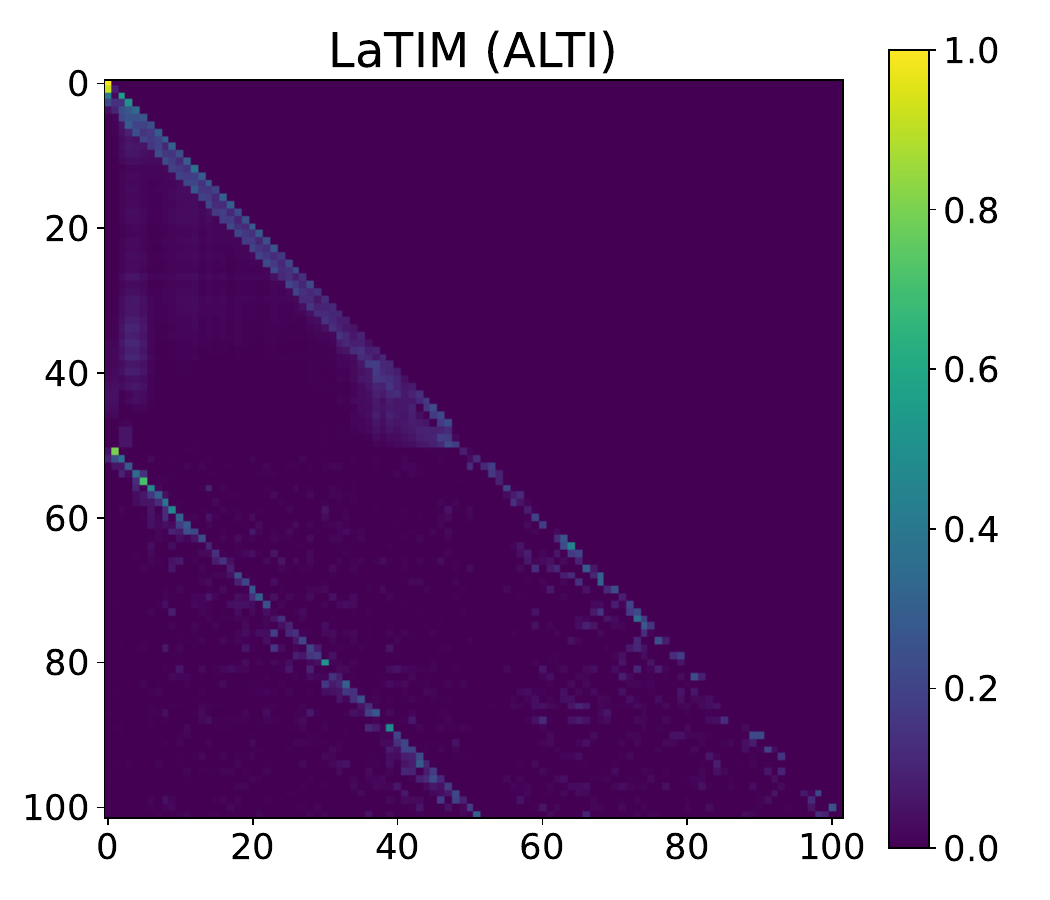}
  \end{subfigure}
  \hfill
  \begin{subfigure}{0.23\textwidth}
    \includegraphics[width=\textwidth]{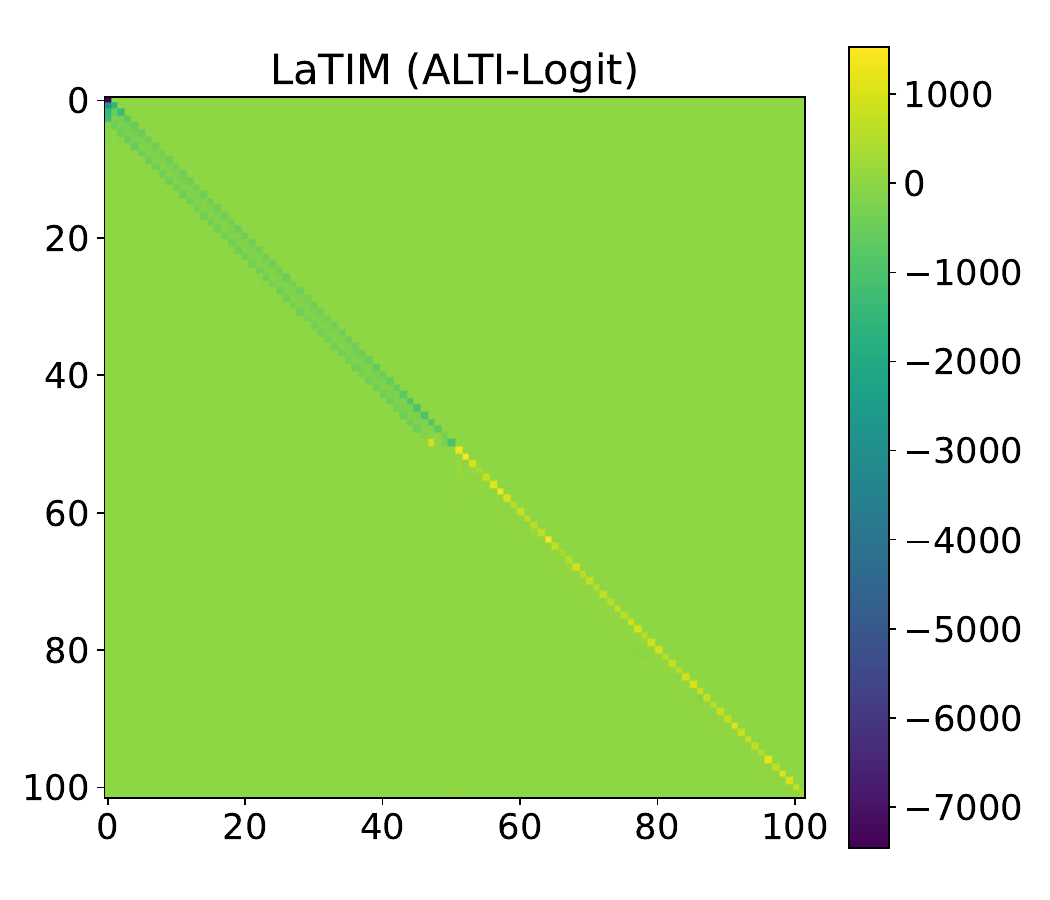}
  \end{subfigure}
  \hfill

 \caption{Heatmaps generated by different interpretability methods for Mamba-2. The interaction between source and copied tokens (along the diagonal line) is more clearly highlighted with \methodname.
 }
  \label{fig:mamba2-copy-alti-vs-avg}
\end{figure}

\section{Experiments}
\label{sec:experiments}

\paragraph{Tasks and Metrics.}
We adopt a diverse set of tasks to provide a rigorous evaluation. 
Following \citet{jelassi2024repeat} 
we experiment on the Copying Task, a synthetic benchmark that tests sequence recall and allows us to faithfully assess how different methods capture token interactions~\citep{bastings-etal-2022-will}. 
Next, we follow \citep{kobayashi-etal-2020-attention} and \citep{ferrando-etal-2022-measuring} and analyze machine translation (MT), where we use the alignment error rate (AER) metric to quantitatively compare the performance of interpretability approaches. 
Finally, we explore retrieval-based generation, leveraging the RULER benchmark \citep{hsieh2024rulerwhatsrealcontext} to investigate Mamba's selective processing in real-world recall-intensive tasks.

\paragraph{Models.}
For machine translation and retrieval-based generation, we use pre-trained versions of Mamba-1 and Mamba-2. 
For the copying task, we train our models from scratch. 
Training details for all tasks are provided in \S\ref{sec:exp-details}.

\paragraph{Methods.}
To evaluate the effectiveness of \methodname, we conduct both qualitative and quantitative assessments, comparing it against existing interpretability techniques for Mamba. 
Namely, we compare our approach against MambaLRP~\citep{jafari2024mambalrp} when using Mamba-1,\footnote{MambaLRP is only defined for Mamba-1.} and with Mamba-Attention/Attribution \citep{ali2024hiddenattentionmambamodels} for both Mamba-1 and Mamba-2. 
Regarding \methodname, we experiment with the variants shown in Table~\ref{tab:equations_overview}.

\subsection{Copying}
\label{sec:copying}

\begin{table}[t]
\centering
\small
\begin{tabular}{lccc}
\toprule
\bf Method & \bf AUC & \bf AP & \bf R@K \\
\midrule
\multicolumn{4}{l}{\textcolor{gray}{\textit{Mamba-1:}}} \\
Mamba-Attention & 0.84 & 0.36 & 0.22 \\ %
Mamba-Attribution & 0.83 & 0.31 & 0.19 \\ %
MambaLRP &  0.40 & 0.22 & 0.20 \\ %
\methodname ($\ell_2$) & \bf 0.88 & 0.41 & 0.27 \\ %
\methodname (ALTI) &  0.86 & \bf 0.47 & \bf 0.36 \\ %
\methodname (ALTI-Logit) & 0.85 & 0.44 & 0.31 \\ %
\cdashlinelr{1-4}
\multicolumn{4}{l}{\textcolor{gray}{\textit{Mamba-2:}}} \\
Mamba-Attention & 0.79 & 0.49 & 0.39 \\ %
Mamba-Attribution &0.79 & 0.47 & 0.39 \\ %
\methodname ($\ell_2$) & \bf 0.98 & \bf 0.86 & \bf 0.74 \\ %
\methodname (ALTI) & 0.85 & 0.71 & 0.63 \\ %
\methodname (ALTI-Logit) & 0.87 & 0.70 & 0.61 \\ %
\bottomrule
\end{tabular}
\normalsize
\caption{Faithfulness evaluation on the copying task in terms of Area Under the Curve (AUC), Average Precision (AP), and Recall at K (R@K).}
\label{tab:copy_faithfulness}
\end{table}

\begin{figure*}[t]
  \centering
  \includegraphics[width=0.245\textwidth]{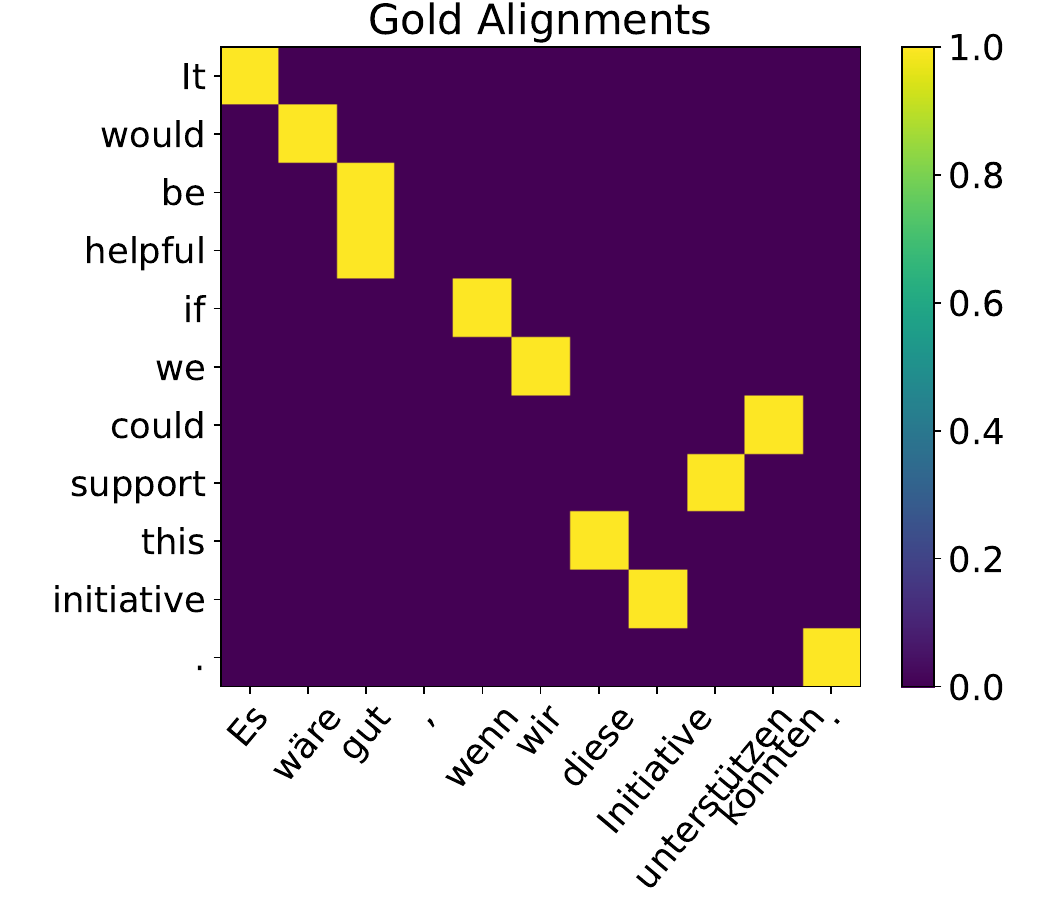}
    \hfill
    \includegraphics[width=0.245\textwidth]{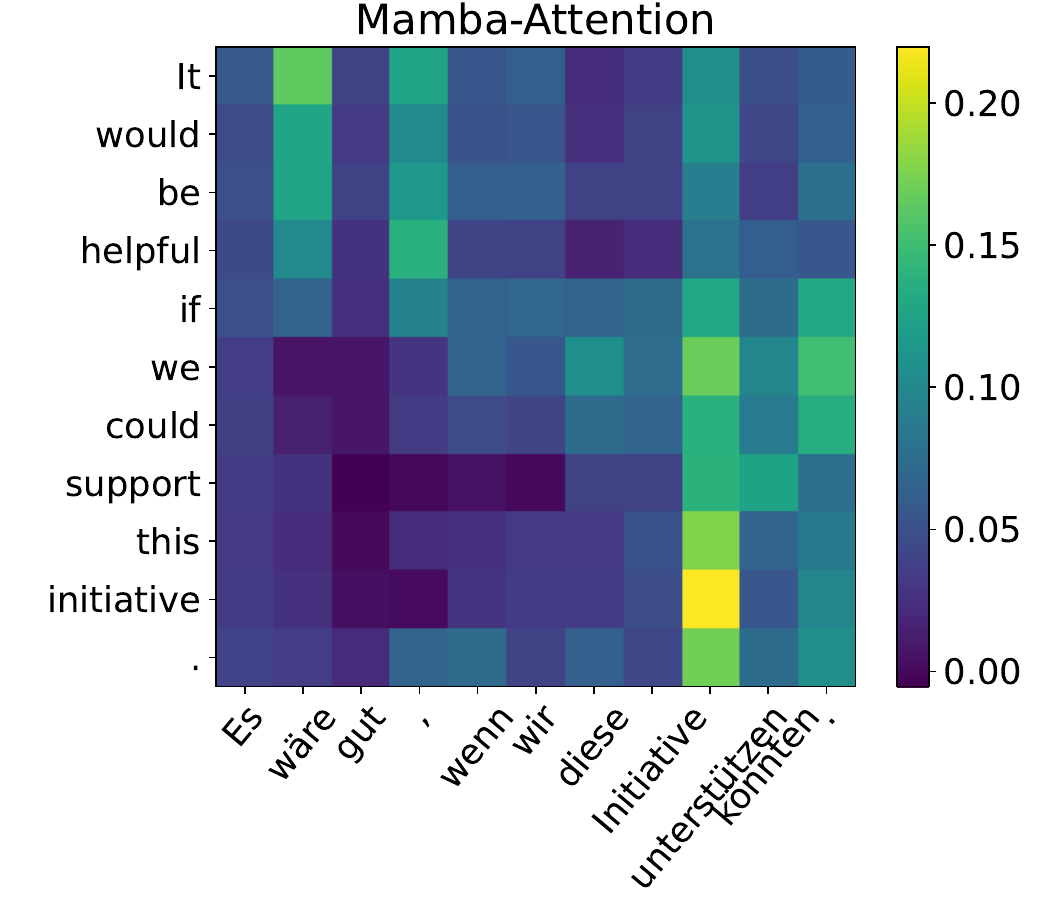}
    \hfill
    \includegraphics[width=0.245\textwidth]{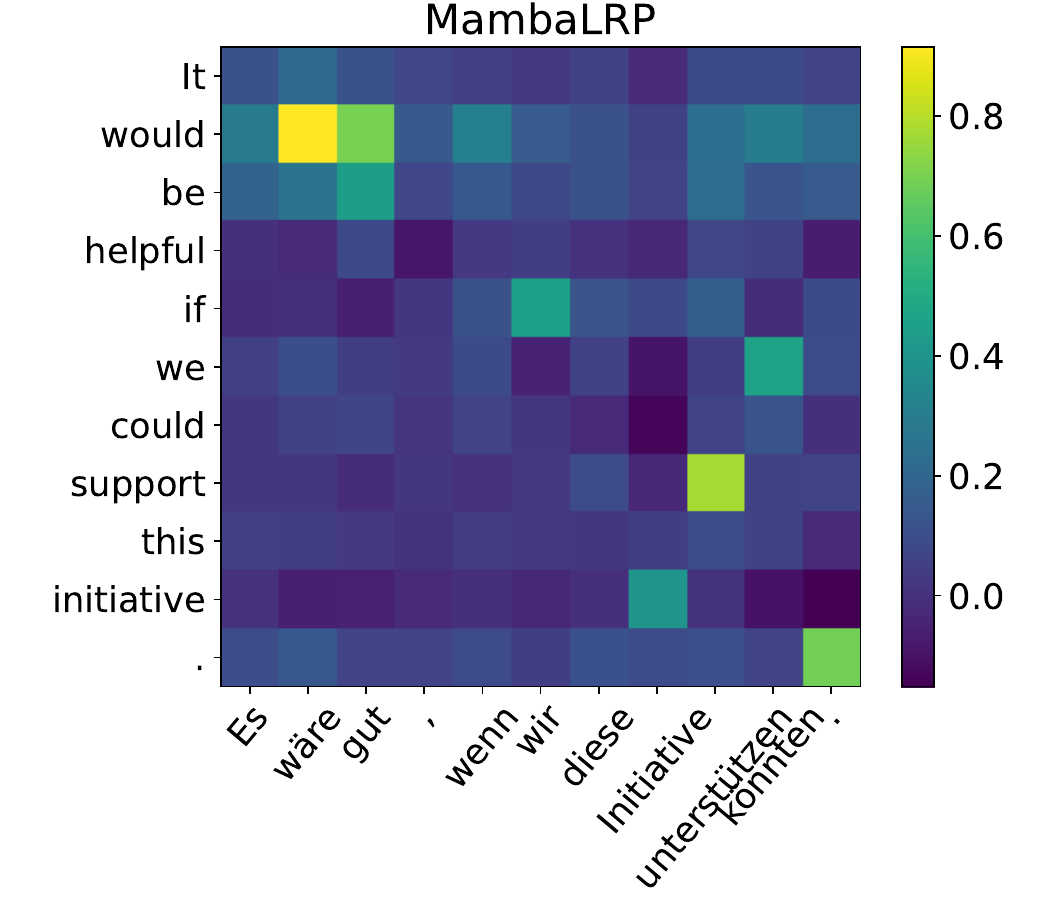}
    \hfill
    \includegraphics[width=0.245\textwidth]{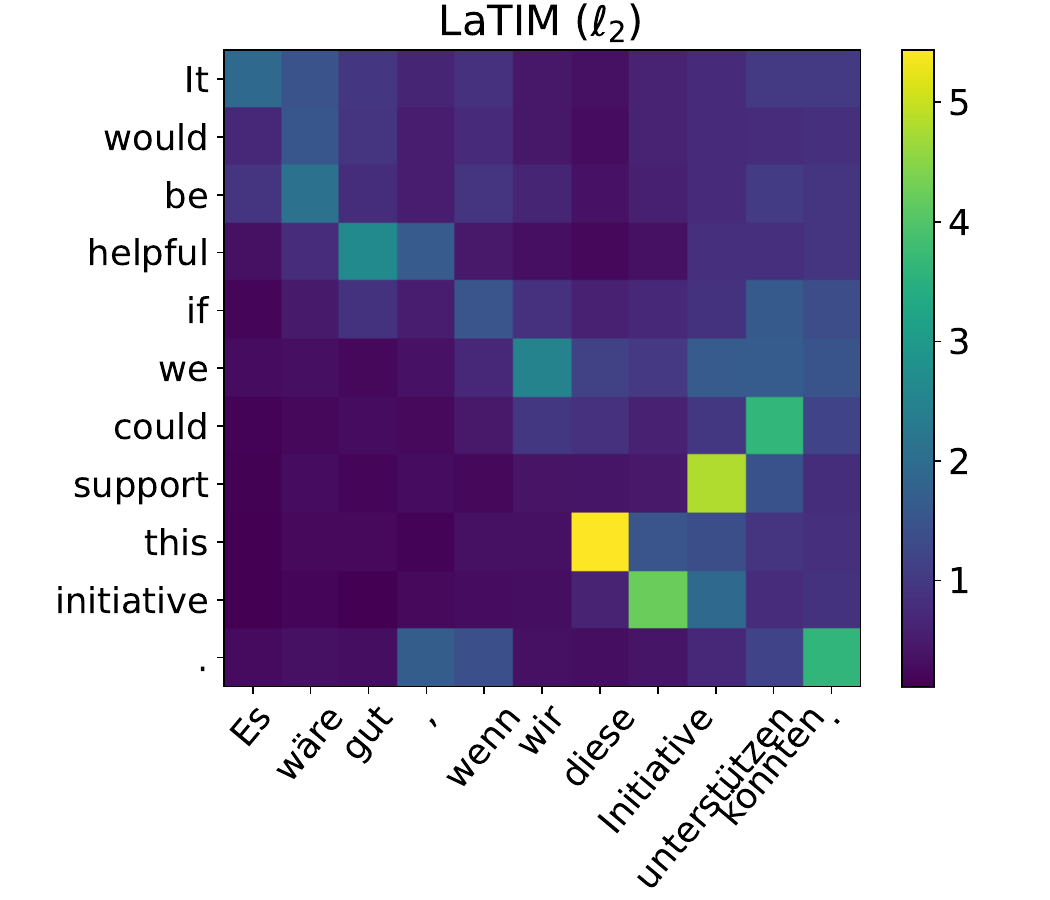}
  \caption{Interpretability heatmaps for Mamba-1 (370M) fine-tuned on \textsc{de$\rightarrow$en} data from the IWSLT17 dataset. 
  \methodname ($\ell_2$) produces alignments that more closely match the ground truth.
  }
  \label{fig:mamba-mt-alti-vs-avg}
\end{figure*}

\begin{table*}[t]
\small
\centering
\setlength{\tabcolsep}{5pt} %
\begin{tabular}{l c c c c c c c c c c c c}
\toprule
 & \multicolumn{4}{c}{\textbf{GoldAlign (\textsc{de\textrightarrow en})}} & \multicolumn{4}{c}{\textbf{IWSLT17 (\textsc{de\textrightarrow en})}} & \multicolumn{4}{c}{\textbf{IWSLT17 (\textsc{fr\textrightarrow en})}} \\
\cmidrule(lr){2-5} \cmidrule(lr){6-9} \cmidrule(lr){10-13}
\textbf{Method} & 
\sc M1\textsubscript{S} & 
\sc M1\textsubscript{L} & 
\sc M2\textsubscript{S} & 
\sc M2\textsubscript{L} & 
\sc M1\textsubscript{S} & 
\sc M1\textsubscript{L} & 
\sc M2\textsubscript{S} & 
\sc M2\textsubscript{L} & 
\sc M1\textsubscript{S} & 
\sc M1\textsubscript{L} & 
\sc M2\textsubscript{S} & 
\sc M2\textsubscript{L} \\  
\midrule
\multicolumn{13}{l}{\textcolor{gray}{\textit{Aggregating layers:}}} \\
MambaLRP                    & 0.50 & 0.47 & - & - & 0.65 & 0.68 & - & - & 0.65 & 0.66 & - & - \\
\methodname (ALTI-Logit)    & 0.68 & 0.69 & 0.63 & 0.69 & 0.67 & 0.71 & 0.60 & 0.74 & 0.71 & 0.69 & 0.62 & 0.76 \\ 
\cdashlinelr{1-13}
\multicolumn{13}{l}{\textcolor{gray}{\textit{Best layer:}}} \\
Mamba-Attention               & 0.84 & 0.85 &  0.84 & 0.85 & 0.79 & 0.79 & 0.72 & 0.79 & 0.80 & 0.79 & 0.69 & 0.78 \\
Mamba-Attribution             & 0.86 & 0.87 & 0.78 & 0.70 & 0.81 & 0.82 & 0.81 & 0.81 & 0.73 & 0.68 & 0.72 & 0.66 \\
\methodname ($\ell_2$)   & \textbf{0.46} & \textbf{0.44} & \textbf{0.49} & 0.52 & \textbf{0.47} & \textbf{0.49} & \textbf{0.43} & 0.49 & \textbf{0.46} & \textbf{0.48} & \textbf{0.35} & \textbf{0.37}  \\
\methodname (ALTI)               & 0.55 & 0.54 & 0.51 & \textbf{0.51} & 0.52 & 0.53 & 0.47 & \textbf{0.47} & 0.53 & 0.53 & 0.38 & 0.38  \\
\bottomrule
\end{tabular}
\normalsize
\caption{Alignment Error Rate (AER) per interpretability method. M1 and M2 stand for Mamba-1 and Mamba-2, with subscript S and M denoting the small (130M) and large (370M) versions, respectively.
}
\label{tab:aer_results}
\end{table*}

The synthetic copying task \citep{jelassi2024repeat} serves as a controlled setting for testing memory recall in SSM-based models, which traditionally struggle with maintaining long-range dependencies \citep{arora2023zoology}. 
Recent advances, such as the mimetic initialization proposed by \citet{trockman2024mimeticinitializationhelpsstate}, have significantly improved Mamba's performance on this task. 
We replicate this setup in a smaller-scale experiment, where 13M models (Mamba-1 and 2) are trained to repeat a 50-token string after a separator token: \texttt{source <SEP> copy}.

\paragraph{Qualitative Analysis.} 
Our interpretability analysis focuses on whether different methods can recover the expected diagonal interaction pattern between source and copied tokens. 
To that end, we start by qualitatively inspecting each method's heatmap in Figure~\ref{fig:mamba2-copy-alti-vs-avg} for Mamba-2.\footnote{We empirically observed that Mamba-1 learns to copy at layer 4, while Mamba-2 shifts this behavior to layer 3. Thus, we extract heatmaps at these layers for the copying task.}
We observe that Mamba-Attention produces a coarse representation of the copy mechanism, lacking the precision needed to capture token-level dependencies. 
In contrast, all \methodname variants better highlight source$\to$copy interactions, making it the superior choice for visualizing the copying mechanism.

\paragraph{Faithfulness Evaluation.} 
To quantitatively assess the reliability of each method, we use a ground-truth matrix with ones along the three main diagonals. 
This means that a faithful interpretability method should produce a well-defined diagonal pattern, indicating that the model correctly attends to preceding tokens, even when shifted, during the copying process.
Leveraging the interpretability metrics from \citet{fomicheva-etal-2021-eval4nlp}, we report a faithfulness evaluation in Table~\ref{tab:copy_faithfulness}.
The results show that all variants of \methodname outperform the baselines, with \methodname ($\ell_2$ and ALTI) consistently achieving the top results across all metrics for both Mamba-1 and Mamba-2.

\begin{figure*}[t]
  \centering
  \includegraphics[width=0.50\textwidth]{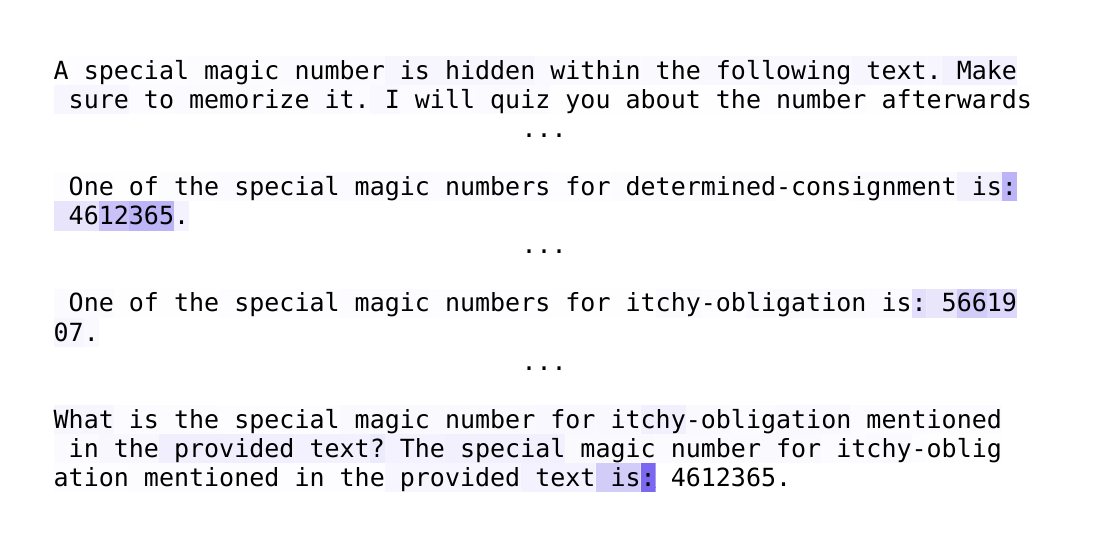}
  \includegraphics[width=0.49\textwidth]{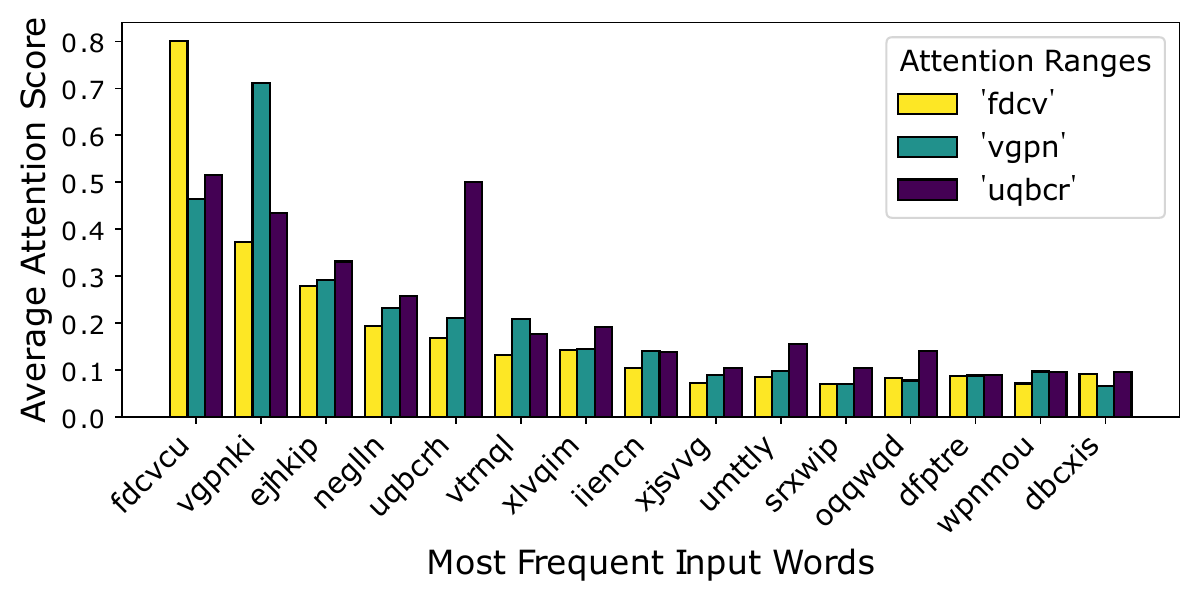}
  \caption{
  Left: 
  Attention map from \methodname ($\ell_2$) for a Passkey Retrieval sample where the key is ``itchy-obligation'' Instead of predicting 5661907, the model incorrectly produces 4612365. 
  Right: Average contribution scores for token ranges preceding each extracted frequent word. 
  Notably, the focus over the token ranges ``fdcv'' and ``vgpn'' aligns well with the two most frequent tokens in the sample (``fdcvcu'', ``vgpnki''). However, when generating ``uqbcr'', it fails to focus on the 3rd most frequent token, suggesting that it relies more on morphological patterns than frequency.\looseness=-1
  }
  \label{fig:mamba-rag}
\end{figure*}

\subsection{Machine Translation}
\label{sec:mt}

We evaluate our method in machine translation (MT) by fine-tuning Mamba models (130M and 370M) on the IWSLT17 dataset \textsc{de$\leftrightarrow$en}~\citep{cettolo2017overview}, following the setup from \citep{pitorro-etal-2024-effective}.
This setup allows us to compare interpretability methods using the alignment error rate (AER), a widely used metric for measuring the accuracy of token alignments in translations. 

\paragraph{Qualitative Analysis.} 
We start by showing the alignments produced by Mamba-1 with the different approaches in Figure~\ref{fig:mamba-mt-alti-vs-avg}, along with the golden alignments provided by \citet{vilar-etal-2006-aer}.
We present additional heatmaps for all methods, including Mamba-2 plots, in Figure~\ref{fig:mamba2-mt-alti-vs-avg} (\S\ref{subsec:app-mt}).
We find that token contribution heatmaps produced by \methodname ($\ell_2$) are sparser and more informative than Mamba-Attention and MambaLRP, which captures the general structure but lacks token-level precision. 
Moreover, we also note that methods that aggregate input relevances across the entire model, such as \methodname (ALTI-Logit), retain sparsity but fail to capture the gold alignments.

\paragraph{Alignment Error Rate.} 
To quantitatively compare methods, we further compute AER on IWSLT17 \textsc{de\textrightarrow en} and \textsc{fr\textrightarrow en} using candidate alignments generated with AwesomeAlign \citep{dou-neubig-2021-word}. 
As seen in Table~\ref{tab:aer_results}, among the layer-wise aggregation methods, we note that MambaLRP consistently outperforms \methodname (ALTI-Logit).
However, when looking at layer-wise methods, we find that \methodname ($\ell_2$) achieves the lowest AER among all methods, reinforcing again its effectiveness in capturing token-to-token interactions, and also suggesting that translation alignments obtained on a per-layer basis might be preferable than those collapsed into a global representation.

\subsection{Retrieval-based Generation}
\label{sec:retrieval_based_generation}

Mamba's efficiency in handling long contexts makes it an attractive candidate for retrieval-based generation. 
However, its ability to selectively recall relevant information remains an open question. 
We investigate this issue using pre-trained Mamba-2 checkpoints with various sizes and experimenting on the RULER benchmark~\citep{hsieh2024rulerwhatsrealcontext}, focusing on two recall-intensive tasks: Passkey Retrieval and Frequent Word Extraction (FWE).

\paragraph{Passkey Retrieval.} In this task, the model must extract a numeric value associated with a key from surrounding distractor text.  In our experiments, Mamba-2 consistently performed well in the simpler, single-passkey setting.
However, as shown in Table~\ref{tab:passkey_acc} (\S\ref{subsec:app-rag}), increasing model size, sequence length, and the number of key-value pairs leads to a significant drop in recall. 
When analyzing attention maps for multi-key retrieval using \methodname ($\ell_2$) in Figure~\ref{fig:mamba-rag} (left), we observe that the 370M model struggles to consistently focus on the correct key, revealing a potential weakness in the multi-key setting.
Additionally, in \S\ref{subsec:app-rag} we 
show that when the gold key appears in the first position, model accuracy declines by 38\% in the 2-key setting and up to 101\% in the 4-key setting. 
Despite these clear limitations, MambaAttention visualization fails to capture this inconsistency, often misattributing focus to misleading tokens.

\paragraph{Frequent Word Extraction.} The FWE task requires the model to extract the three most frequent synthetic words in a passage. 
In \S\ref{subsec:app-rag}, we show that Mamba models, even at the 1.4B parameter scale, struggle with this task.
Our analysis in Figure~\ref{fig:mamba-rag} (right), using \methodname ($\ell_2$), reveals that the model frequently misidentifies the correct 3rd most frequent token, highlighting its difficulty in tracking long-range token occurrences. 
We also note that Mamba's attention on repeated words decays over time, 
which may explain its failure to accurately count word frequency.

\subsection{Approximation Error Analysis}
\label{subsec:approximation_error_analysis}

As noted in Section~\ref{subsec:decomposition_error}, our current method involves an approximate decomposition of Mamba's computations due to the non-linearity introduced by the $\textsf{SiLU}$ activation.
To measure the impact of this approximation, we experiment with alternative activations by retraining Mamba with ReLU or disabling activations entirely, which casts $f$ as the identity function and, more importantly, yields an \emph{exact method}. 
We perform continued pretraining of Mamba-2 (370M) on the FineWeb-Edu dataset~\citep{penedo2024the} and evaluate on the IWSLT17 \textsc{de$\to$en} dataset using AER to assess interpretability and COMET~\citep{rei-etal-2020-comet} to asses translation quality.
Results are shown in Table~\ref{tab:mt_approx_error}.
Interestingly, a model trained without a non-linear activation achieves not only 0 approximation error but also leads to the best AER scores along with a high COMET. 
As noted by \citet{bick2024transformers}, who also disable the activation before SSM distillation, a purely linear variant of Mamba can be an effective alternative for more interpretable architectures. 
Nonetheless, we highlight that our approximated version with $f := \textsf{SiLU}$ leads to similar AER and COMET scores as $f := \textsf{identity}$.

\section{Related Work}

\paragraph{Input Attribution Methods.}
A large body of work focuses on interpretability via input attribution, particularly in transformers, where attention maps serve as a widely used technique~\citep{fantozzi2024explainability}. While attention weights alone can be unfaithful indicators of model decisions~\citep{jain-wallace-2019-attention,bastings-filippova-2020-elephant}, they remain useful in many applications, including machine translation~\citep{wiegreffe-pinter-2019-attention,treviso-martins-2020-explanation}. 
Recent methods go beyond simple attention analysis by explicitly decomposing internal model computations, such as integrating value-weighted norms~\citep{kobayashi-etal-2020-attention} or using vector distances to estimate token contributions~\citep{ferrando-etal-2022-measuring}. 
Additionally, aggregation-based techniques, including Attention Rollout~\citep{abnar-zuidema-2020-quantifying}, DiffMask~\citep{de-cao-etal-2020-decisions}, and ALTI-Logit~\citep{ferrando-etal-2023-explaining}, consolidate relevance scores across multiple layers to provide a more holistic view of information flow.
While these methods have substantially improved transformer interpretability, state space models (SSMs) remain comparatively underexplored.

\paragraph{Theoretical Insights into SSMs.}
Beyond interpretability, several studies have analyzed the internal mechanisms of SSMs. 
\citet{vo2025demystifying} investigate the asymptotic behavior of token states, revealing conditions under which tokens either converge or diverge, affecting memory retention.
\citet{sieber2024understanding} introduce a framework that unifies different sequence modeling paradigms, including SSMs, under a common mathematical representation.
Meanwhile, \citet{trockman2024mimeticinitializationhelpsstate} propose an initialization technique that improves Mamba's recall ability inspired by attention-like patterns.

\paragraph{Interpreting Mamba.}
Despite the growing adoption of Mamba, only a few works have explicitly addressed its interpretability.
\citet{ali2024hiddenattentionmambamodels} introduce Mamba-Attention and Mamba-Attribution, which approximate token interactions by extracting implicit attention patterns in Mamba-1.
Similarly, MambaLRP~\citep{jafari2024mambalrp} applies Layer-wise Relevance Propagation to Mamba-1, ensuring stable attribution propagation.
However, these approaches do not provide a direct decomposition of individual token contributions, leaving gaps in understanding how Mamba selectively processes information.
\methodname bridges this gap by providing fine-grained, token-level interpretability for both Mamba-1 and Mamba-2. Additionally, we note that \methodname is adaptable and can be applied to other linear recurrent architectures, such as DeltaNet~\citep{yang2024parallelizing} and mLSTM~\citep{beck2024xlstm}, making it a valuable interpretability tool for long-context models.

\begin{table}[t]
\centering
\small
\setlength{\tabcolsep}{4.5pt} %
\begin{tabular}{lccccc}
\toprule
\multirow{2}{*}{\textbf{Activation}} & \multicolumn{3}{c}{\textbf{Error per Layer}} & \multirow{2}{*}{\textbf{AER}} & \multirow{2}{*}{\textbf{COMET}}\\
\cmidrule(lr){2-4}
& 0-16 & 16-32 & 32-48 & \\
\midrule
\textsf{SiLU} & 0.21  & 0.45  & 0.57 & 0.47 & 83.4 \\
\textsf{SiLU} + CP & 0.21 & 0.43 & 0.54 & \textbf{0.46} & \textbf{83.6} \\
\textsf{ReLU} & 0.35 & 0.83 & 1.07 & 0.51 & 82.8 \\
\textsf{Identity} & \textbf{0.00} & \textbf{0.00}  & \textbf{0.00} & \textbf{0.46} & 83.3 \\
\bottomrule
\end{tabular}
\caption{Approximation error analysis with different activations for computing $\bm{\phi}_i$ in Equation~\ref{eq:silu_placement}. 
CP indicates continued pretraining. }
\label{tab:mt_approx_error}
\end{table}

\section{Conclusion}
\label{sec:conclusion}

Our experiments demonstrate that our token-level decomposition approach significantly improves interpretability for Mamba models. Across copying, machine translation, and retrieval-based generation tasks, we show that our method, \methodname, provides clearer insights into Mamba's selective processing mechanisms. 
For example, our findings suggest that Mamba's recall limitations in long-context tasks may stem from its sparse and decaying focus on relevant tokens. 
Moreover, our study confirms that while \methodname introduces a minimal approximation error, its exact counterpart eliminates this error entirely while maintaining interpretability and task performance.
Together, these contributions improve our understanding of Mamba and open new directions for improving its reliability and effectiveness in real-world applications.

\section*{Acknowledgements}
We thank André Martins, Pavlo Vasylenko, Giuseppe Attanasio and Saúl Santos for their helpful and constructive feedback. This work was supported by EU's Horizon Europe Research and Innovation Actions (UTTER, contract 101070631), by the project DECOLLAGE (ERC-2022-CoG 101088763), by the Portuguese Recovery and Resilience Plan through project C645008882-00000055 (Center for Responsible AI), and by FCT/MECI through national funds and when applicable co-funded EU funds under UID/50008: Instituto de Telecomunicações.

\section*{Limitations}

We point out some limitations of the presented
study.
Our method, \methodname, relies on an approximation strategy to decompose token contributions due to the non-linearity introduced by the \textsf{SiLU} activation. Although our empirical analysis suggests that this approximation does not meaningfully impact interpretability quality, an exact decomposition requires model modifications, such as removing non-linearities, requiring re-training. 
Additionally, our evaluation focuses primarily on tasks like Copying and Machine Translation, where token interactions are well understood. In more complex tasks such as Retrieval-based Generation, assessing interpretability quality is harder, and further validation with human evaluations could provide a more robust assessment.

Furthermore, \methodname is specifically designed for Mamba-1 and Mamba-2, and while the principles behind it could easily be extended to other state space models or linear recurrent models, some additional modifications may be necessary. 
Architectures incorporating more complex gating mechanisms or hybrid attention-SSM layers might require adapted decomposition techniques. Additionally, while \methodname helps visualize token interactions, its impact on improving model robustness and trustworthiness remains an open question.

\section*{Potential Risks}
Although our token-level decomposition provides valuable insights, it may also be misused. An overreliance on the generated token maps could lead users to assume these partial explanations capture all aspects of the model's reasoning. This false confidence may mask biases in the model or data, and encourage trust in outputs without adequate scrutiny, particularly in sensitive domains.

Additionally, exposing how Mamba selectively processes tokens could aid malicious actors in crafting targeted adversarial inputs. By identifying which tokens or positions most influence the model, adversaries could exploit these patterns to degrade performance or manipulate outputs. Such misuse risks undermining the reliability of Mamba-based systems, especially when high-stakes decisions rely on accurate and fair model predictions.

\bibliography{anthology_small, custom}

\begin{thebibliography}{46}
\providecommand{\natexlab}[1]{#1}

\bibitem[{Abnar and Zuidema(2020)}]{abnar-zuidema-2020-quantifying}
Samira Abnar and Willem Zuidema. 2020.
\newblock \href {https://doi.org/10.18653/v1/2020.acl-main.385} {Quantifying attention flow in transformers}.
\newblock In \emph{Proceedings of the 58th Annual Meeting of the Association for Computational Linguistics}, pages 4190--4197, Online. Association for Computational Linguistics.

\bibitem[{Ali et~al.(2024)Ali, Zimerman, and Wolf}]{ali2024hiddenattentionmambamodels}
Ameen Ali, Itamar Zimerman, and Lior Wolf. 2024.
\newblock \href {https://arxiv.org/abs/2403.01590} {The hidden attention of mamba models}.
\newblock \emph{Preprint}, arXiv:2403.01590.

\bibitem[{Arora et~al.(2024)Arora, Eyuboglu, Timalsina, Johnson, Poli, Zou, Rudra, and Re}]{arora2023zoology}
Simran Arora, Sabri Eyuboglu, Aman Timalsina, Isys Johnson, Michael Poli, James Zou, Atri Rudra, and Christopher Re. 2024.
\newblock \href {https://openreview.net/forum?id=LY3ukUANko} {Zoology: Measuring and improving recall in efficient language models}.
\newblock In \emph{The Twelfth International Conference on Learning Representations}.

\bibitem[{Ba et~al.(2016)Ba, Kiros, and Hinton}]{ba2016layer}
Jimmy~Lei Ba, Jamie~Ryan Kiros, and Geoffrey~E. Hinton. 2016.
\newblock \href {https://arxiv.org/abs/1607.06450} {Layer normalization}.
\newblock \emph{Preprint}, arXiv:1607.06450.

\bibitem[{Bastings et~al.(2022)Bastings, Ebert, Zablotskaia, Sandholm, and Filippova}]{bastings-etal-2022-will}
Jasmijn Bastings, Sebastian Ebert, Polina Zablotskaia, Anders Sandholm, and Katja Filippova. 2022.
\newblock \href {https://doi.org/10.18653/v1/2022.emnlp-main.64} {{``}will you find these shortcuts?{''} a protocol for evaluating the faithfulness of input salience methods for text classification}.
\newblock In \emph{Proceedings of the 2022 Conference on Empirical Methods in Natural Language Processing}, pages 976--991, Abu Dhabi, United Arab Emirates. Association for Computational Linguistics.

\bibitem[{Bastings and Filippova(2020)}]{bastings-filippova-2020-elephant}
Jasmijn Bastings and Katja Filippova. 2020.
\newblock \href {https://doi.org/10.18653/v1/2020.blackboxnlp-1.14} {The elephant in the interpretability room: Why use attention as explanation when we have saliency methods?}
\newblock In \emph{Proceedings of the Third BlackboxNLP Workshop on Analyzing and Interpreting Neural Networks for NLP}, pages 149--155, Online. Association for Computational Linguistics.

\bibitem[{Beck et~al.(2024)Beck, P{\"o}ppel, Spanring, Auer, Prudnikova, Kopp, Klambauer, Brandstetter, and Hochreiter}]{beck2024xlstm}
Maximilian Beck, Korbinian P{\"o}ppel, Markus Spanring, Andreas Auer, Oleksandra Prudnikova, Michael~K Kopp, G{\"u}nter Klambauer, Johannes Brandstetter, and Sepp Hochreiter. 2024.
\newblock \href {https://openreview.net/forum?id=ARAxPPIAhq} {x{LSTM}: Extended long short-term memory}.
\newblock In \emph{The Thirty-eighth Annual Conference on Neural Information Processing Systems}.

\bibitem[{Bick et~al.(2024)Bick, Li, Xing, Kolter, and Gu}]{bick2024transformers}
Aviv Bick, Kevin Li, Eric~P. Xing, J~Zico Kolter, and Albert Gu. 2024.
\newblock \href {https://openreview.net/forum?id=FJlrSZBMCD} {Transformers to {SSM}s: Distilling quadratic knowledge to subquadratic models}.
\newblock In \emph{The Thirty-eighth Annual Conference on Neural Information Processing Systems}.

\bibitem[{Cettolo et~al.(2017{\natexlab{a}})Cettolo, Federico, Bentivogli, Jan, Sebastian, Katsuitho, Koichiro, and Christian}]{cettolo2017overview}
Mauro Cettolo, Marcello Federico, Luisa Bentivogli, Niehues Jan, St{\"u}ker Sebastian, Sudoh Katsuitho, Yoshino Koichiro, and Federmann Christian. 2017{\natexlab{a}}.
\newblock Overview of the iwslt 2017 evaluation campaign.
\newblock In \emph{Proceedings of the 14th International Workshop on Spoken Language Translation (IWSLT)}, pages 2--14.

\bibitem[{Cettolo et~al.(2017{\natexlab{b}})Cettolo, Federico, Bentivogli, Niehues, St{\"u}ker, Sudoh, Yoshino, and Federmann}]{cettolo-etal-2017-overview}
Mauro Cettolo, Marcello Federico, Luisa Bentivogli, Jan Niehues, Sebastian St{\"u}ker, Katsuhito Sudoh, Koichiro Yoshino, and Christian Federmann. 2017{\natexlab{b}}.
\newblock \href {https://aclanthology.org/2017.iwslt-1.1} {Overview of the {IWSLT} 2017 evaluation campaign}.
\newblock In \emph{Proceedings of the 14th International Conference on Spoken Language Translation}, pages 2--14, Tokyo, Japan. International Workshop on Spoken Language Translation.

\bibitem[{Dao and Gu(2024)}]{dao2024transformersssmsgeneralizedmodels}
Tri Dao and Albert Gu. 2024.
\newblock \href {https://arxiv.org/abs/2405.21060} {Transformers are ssms: Generalized models and efficient algorithms through structured state space duality}.
\newblock \emph{Preprint}, arXiv:2405.21060.

\bibitem[{De~Cao et~al.(2020)De~Cao, Schlichtkrull, Aziz, and Titov}]{de-cao-etal-2020-decisions}
Nicola De~Cao, Michael~Sejr Schlichtkrull, Wilker Aziz, and Ivan Titov. 2020.
\newblock \href {https://doi.org/10.18653/v1/2020.emnlp-main.262} {How do decisions emerge across layers in neural models? interpretation with differentiable masking}.
\newblock In \emph{Proceedings of the 2020 Conference on Empirical Methods in Natural Language Processing (EMNLP)}, pages 3243--3255, Online. Association for Computational Linguistics.

\bibitem[{Dong et~al.(2025)Dong, Fu, Diao, Byeon, Chen, Mahabaleshwarkar, Liu, keirsbilck, Chen, Suhara, Lin, Kautz, and Molchanov}]{dong2025hymba}
Xin Dong, Yonggan Fu, Shizhe Diao, Wonmin Byeon, Zijia Chen, Ameya~Sunil Mahabaleshwarkar, Shih-Yang Liu, Matthijs~Van keirsbilck, Min-Hung Chen, Yoshi Suhara, Yingyan~Celine Lin, Jan Kautz, and Pavlo Molchanov. 2025.
\newblock \href {https://openreview.net/forum?id=A1ztozypga} {Hymba: A hybrid-head architecture for small language models}.
\newblock In \emph{The Thirteenth International Conference on Learning Representations}.

\bibitem[{Dou and Neubig(2021)}]{dou-neubig-2021-word}
Zi-Yi Dou and Graham Neubig. 2021.
\newblock \href {https://doi.org/10.18653/v1/2021.eacl-main.181} {Word alignment by fine-tuning embeddings on parallel corpora}.
\newblock In \emph{Proceedings of the 16th Conference of the European Chapter of the Association for Computational Linguistics: Main Volume}, pages 2112--2128, Online. Association for Computational Linguistics.

\bibitem[{Fantozzi and Naldi(2024)}]{fantozzi2024explainability}
Paolo Fantozzi and Maurizio Naldi. 2024.
\newblock The explainability of transformers: Current status and directions.
\newblock \emph{Computers}, 13(4):92.

\bibitem[{Ferrando et~al.(2022)Ferrando, G{\'a}llego, and Costa-juss{\`a}}]{ferrando-etal-2022-measuring}
Javier Ferrando, Gerard~I. G{\'a}llego, and Marta~R. Costa-juss{\`a}. 2022.
\newblock \href {https://doi.org/10.18653/v1/2022.emnlp-main.595} {Measuring the mixing of contextual information in the transformer}.
\newblock In \emph{Proceedings of the 2022 Conference on Empirical Methods in Natural Language Processing}, pages 8698--8714, Abu Dhabi, United Arab Emirates. Association for Computational Linguistics.

\bibitem[{Ferrando et~al.(2023)Ferrando, G{\'a}llego, Tsiamas, and Costa-juss{\`a}}]{ferrando-etal-2023-explaining}
Javier Ferrando, Gerard~I. G{\'a}llego, Ioannis Tsiamas, and Marta~R. Costa-juss{\`a}. 2023.
\newblock \href {https://doi.org/10.18653/v1/2023.acl-long.301} {Explaining how transformers use context to build predictions}.
\newblock In \emph{Proceedings of the 61st Annual Meeting of the Association for Computational Linguistics (Volume 1: Long Papers)}, pages 5486--5513, Toronto, Canada. Association for Computational Linguistics.

\bibitem[{Ferrando et~al.(2024)Ferrando, Sarti, Bisazza, and Costa-juss{\`a}}]{ferrando-etal-2024-primer}
Javier Ferrando, Gabriele Sarti, Arianna Bisazza, and Marta~R Costa-juss{\`a}. 2024.
\newblock \href {http://arxiv.org/abs/2405.00208} {A primer on the inner workings of transformer-based language models}.
\newblock \emph{arXiv preprint arXiv:2405.00208}.

\bibitem[{Fomicheva et~al.(2021)Fomicheva, Lertvittayakumjorn, Zhao, Eger, and Gao}]{fomicheva-etal-2021-eval4nlp}
Marina Fomicheva, Piyawat Lertvittayakumjorn, Wei Zhao, Steffen Eger, and Yang Gao. 2021.
\newblock \href {https://doi.org/10.18653/v1/2021.eval4nlp-1.17} {The {E}val4{NLP} shared task on explainable quality estimation: Overview and results}.
\newblock In \emph{Proceedings of the 2nd Workshop on Evaluation and Comparison of NLP Systems}, pages 165--178, Punta Cana, Dominican Republic. Association for Computational Linguistics.

\bibitem[{Gu and Dao(2023)}]{gu2023mamba}
Albert Gu and Tri Dao. 2023.
\newblock \href {https://arxiv.org/abs/2312.00752} {Mamba: Linear-time sequence modeling with selective state spaces}.
\newblock \emph{Preprint}, arXiv:2312.00752.

\bibitem[{Gu et~al.(2020)Gu, Dao, Ermon, Rudra, and R\'{e}}]{gu2020hippo}
Albert Gu, Tri Dao, Stefano Ermon, Atri Rudra, and Christopher R\'{e}. 2020.
\newblock \href {https://proceedings.neurips.cc/paper_files/paper/2020/file/102f0bb6efb3a6128a3c750dd16729be-Paper.pdf} {Hippo: Recurrent memory with optimal polynomial projections}.
\newblock In \emph{Advances in Neural Information Processing Systems}, volume~33, pages 1474--1487. Curran Associates, Inc.

\bibitem[{Gu et~al.(2022)Gu, Goel, and Re}]{gu2022efficiently}
Albert Gu, Karan Goel, and Christopher Re. 2022.
\newblock \href {https://openreview.net/forum?id=uYLFoz1vlAC} {Efficiently modeling long sequences with structured state spaces}.
\newblock In \emph{International Conference on Learning Representations}.

\bibitem[{Hsieh et~al.(2024)Hsieh, Sun, Kriman, Acharya, Rekesh, Jia, and Ginsburg}]{hsieh2024rulerwhatsrealcontext}
Cheng-Ping Hsieh, Simeng Sun, Samuel Kriman, Shantanu Acharya, Dima Rekesh, Fei Jia, and Boris Ginsburg. 2024.
\newblock \href {https://openreview.net/forum?id=kIoBbc76Sy} {{RULER}: What{\textquoteright}s the real context size of your long-context language models?}
\newblock In \emph{First Conference on Language Modeling}.

\bibitem[{Hu et~al.(2024)Hu, Tu, Han, He, Cui, Long, Zheng, Fang, Huang, Zhao, Zhang, Thai, Zhang, Wang, Yao, Zhao, Zhou, Cai, Zhai, Ding, Jia, Zeng, Li, Liu, and Sun}]{hu2024minicpmunveilingpotentialsmall}
Shengding Hu, Yuge Tu, Xu~Han, Chaoqun He, Ganqu Cui, Xiang Long, Zhi Zheng, Yewei Fang, Yuxiang Huang, Weilin Zhao, Xinrong Zhang, Zheng~Leng Thai, Kaihuo Zhang, Chongyi Wang, Yuan Yao, Chenyang Zhao, Jie Zhou, Jie Cai, Zhongwu Zhai, Ning Ding, Chao Jia, Guoyang Zeng, Dahai Li, Zhiyuan Liu, and Maosong Sun. 2024.
\newblock \href {https://arxiv.org/abs/2404.06395} {Minicpm: Unveiling the potential of small language models with scalable training strategies}.
\newblock \emph{Preprint}, arXiv:2404.06395.

\bibitem[{Jafari et~al.(2024)Jafari, Montavon, Muller, and Eberle}]{jafari2024mambalrp}
Farnoush~Rezaei Jafari, Gr{\'e}goire Montavon, Klaus~Robert Muller, and Oliver Eberle. 2024.
\newblock \href {https://openreview.net/forum?id=2n1Ysn1EDl} {Mamba{LRP}: Explaining selective state space sequence models}.
\newblock In \emph{The Thirty-eighth Annual Conference on Neural Information Processing Systems}.

\bibitem[{Jain and Wallace(2019)}]{jain-wallace-2019-attention}
Sarthak Jain and Byron~C. Wallace. 2019.
\newblock \href {https://doi.org/10.18653/v1/N19-1357} {{A}ttention is not {E}xplanation}.
\newblock In \emph{Proceedings of the 2019 Conference of the North {A}merican Chapter of the Association for Computational Linguistics: Human Language Technologies, Volume 1 (Long and Short Papers)}, pages 3543--3556, Minneapolis, Minnesota. Association for Computational Linguistics.

\bibitem[{Jelassi et~al.(2024)Jelassi, Brandfonbrener, Kakade, and eran malach}]{jelassi2024repeat}
Samy Jelassi, David Brandfonbrener, Sham~M. Kakade, and eran malach. 2024.
\newblock \href {https://openreview.net/forum?id=duRRoGeoQT} {Repeat after me: Transformers are better than state space models at copying}.
\newblock In \emph{Forty-first International Conference on Machine Learning}.

\bibitem[{Kobayashi et~al.(2020)Kobayashi, Kuribayashi, Yokoi, and Inui}]{kobayashi-etal-2020-attention}
Goro Kobayashi, Tatsuki Kuribayashi, Sho Yokoi, and Kentaro Inui. 2020.
\newblock \href {https://doi.org/10.18653/v1/2020.emnlp-main.574} {Attention is not only a weight: Analyzing transformers with vector norms}.
\newblock In \emph{Proceedings of the 2020 Conference on Empirical Methods in Natural Language Processing (EMNLP)}, pages 7057--7075, Online. Association for Computational Linguistics.

\bibitem[{Kobayashi et~al.(2021)Kobayashi, Kuribayashi, Yokoi, and Inui}]{kobayashi-etal-2021-incorporating}
Goro Kobayashi, Tatsuki Kuribayashi, Sho Yokoi, and Kentaro Inui. 2021.
\newblock \href {https://doi.org/10.18653/v1/2021.emnlp-main.373} {{I}ncorporating {R}esidual and {N}ormalization {L}ayers into {A}nalysis of {M}asked {L}anguage {M}odels}.
\newblock In \emph{Proceedings of the 2021 Conference on Empirical Methods in Natural Language Processing}, pages 4547--4568, Online and Punta Cana, Dominican Republic. Association for Computational Linguistics.

\bibitem[{Lenz et~al.(2025)Lenz, Lieber, Arazi, Bergman, Manevich, Peleg, Aviram, Almagor, Fridman, Padnos, Gissin, Jannai, Muhlgay, Zimberg, Gerber, Dolev, Krakovsky, Safahi, Schwartz, Cohen, Shachaf, Rozenblum, Bata, Blass, Magar, Dalmedigos, Osin, Fadlon, Rozman, Danos, Gokhman, Zusman, Gidron, Ratner, Gat, Rozen, Fried, Leshno, Antverg, Abend, Dagan, Cohavi, Alon, Belson, Cohen, Gilad, Glozman, Lev, Shalev-Shwartz, Meirom, Delbari, Ness, Asida, Gal, Braude, Pumerantz, Cohen, Belinkov, Globerson, Levy, and Shoham}]{lenz2025jamba}
Barak Lenz, Opher Lieber, Alan Arazi, Amir Bergman, Avshalom Manevich, Barak Peleg, Ben Aviram, Chen Almagor, Clara Fridman, Dan Padnos, Daniel Gissin, Daniel Jannai, Dor Muhlgay, Dor Zimberg, Edden~M. Gerber, Elad Dolev, Eran Krakovsky, Erez Safahi, Erez Schwartz, Gal Cohen, Gal Shachaf, Haim Rozenblum, Hofit Bata, Ido Blass, Inbal Magar, Itay Dalmedigos, Jhonathan Osin, Julie Fadlon, Maria Rozman, Matan Danos, Michael Gokhman, Mor Zusman, Naama Gidron, Nir Ratner, Noam Gat, Noam Rozen, Oded Fried, Ohad Leshno, Omer Antverg, Omri Abend, Or~Dagan, Orit Cohavi, Raz Alon, Ro'i Belson, Roi Cohen, Rom Gilad, Roman Glozman, Shahar Lev, Shai Shalev-Shwartz, Shaked~Haim Meirom, Tal Delbari, Tal Ness, Tomer Asida, Tom~Ben Gal, Tom Braude, Uriya Pumerantz, Josh Cohen, Yonatan Belinkov, Yuval Globerson, Yuval~Peleg Levy, and Yoav Shoham. 2025.
\newblock \href {https://openreview.net/forum?id=JFPaD7lpBD} {Jamba: Hybrid transformer-mamba language models}.
\newblock In \emph{The Thirteenth International Conference on Learning Representations}.

\bibitem[{Loshchilov and Hutter(2019)}]{loshchilov2019decoupled}
Ilya Loshchilov and Frank Hutter. 2019.
\newblock \href {https://openreview.net/forum?id=Bkg6RiCqY7} {Decoupled weight decay regularization}.
\newblock In \emph{International Conference on Learning Representations}.

\bibitem[{Mohebbi et~al.(2024)Mohebbi, Jumelet, Hanna, Alishahi, and Zuidema}]{mohebbi-etal-2024-transformer}
Hosein Mohebbi, Jaap Jumelet, Michael Hanna, Afra Alishahi, and Willem Zuidema. 2024.
\newblock \href {https://aclanthology.org/2024.eacl-tutorials.4} {Transformer-specific interpretability}.
\newblock In \emph{Proceedings of the 18th Conference of the European Chapter of the Association for Computational Linguistics: Tutorial Abstracts}, pages 21--26, St. Julian{'}s, Malta. Association for Computational Linguistics.

\bibitem[{Penedo et~al.(2024)Penedo, Kydl{\'\i}{\v{c}}ek, allal, Lozhkov, Mitchell, Raffel, Werra, and Wolf}]{penedo2024the}
Guilherme Penedo, Hynek Kydl{\'\i}{\v{c}}ek, Loubna~Ben allal, Anton Lozhkov, Margaret Mitchell, Colin Raffel, Leandro~Von Werra, and Thomas Wolf. 2024.
\newblock \href {https://openreview.net/forum?id=n6SCkn2QaG} {The fineweb datasets: Decanting the web for the finest text data at scale}.
\newblock In \emph{The Thirty-eight Conference on Neural Information Processing Systems Datasets and Benchmarks Track}.

\bibitem[{Pitorro et~al.(2024)Pitorro, Vasylenko, Treviso, and Martins}]{pitorro-etal-2024-effective}
Hugo Pitorro, Pavlo Vasylenko, Marcos Treviso, and Andr{\'e} Martins. 2024.
\newblock \href {https://aclanthology.org/2024.wmt-1.111} {How effective are state space models for machine translation?}
\newblock In \emph{Proceedings of the Ninth Conference on Machine Translation}, pages 1107--1124, Miami, Florida, USA. Association for Computational Linguistics.

\bibitem[{Rei et~al.(2020)Rei, Stewart, Farinha, and Lavie}]{rei-etal-2020-comet}
Ricardo Rei, Craig Stewart, Ana~C Farinha, and Alon Lavie. 2020.
\newblock \href {https://doi.org/10.18653/v1/2020.emnlp-main.213} {{COMET}: A neural framework for {MT} evaluation}.
\newblock In \emph{Proceedings of the 2020 Conference on Empirical Methods in Natural Language Processing (EMNLP)}, pages 2685--2702, Online. Association for Computational Linguistics.

\bibitem[{Sieber et~al.(2024)Sieber, Alonso, Didier, Zeilinger, and Orvieto}]{sieber2024understanding}
Jerome Sieber, Carmen~Amo Alonso, Alexandre Didier, Melanie Zeilinger, and Antonio Orvieto. 2024.
\newblock \href {https://openreview.net/forum?id=iF7MnXnxRw} {Understanding the differences in foundation models: Attention, state space models, and recurrent neural networks}.
\newblock In \emph{The Thirty-eighth Annual Conference on Neural Information Processing Systems}.

\bibitem[{Srivastava et~al.(2014)Srivastava, Hinton, Krizhevsky, Sutskever, and Salakhutdinov}]{JMLR:v15:srivastava14a}
Nitish Srivastava, Geoffrey Hinton, Alex Krizhevsky, Ilya Sutskever, and Ruslan Salakhutdinov. 2014.
\newblock \href {http://jmlr.org/papers/v15/srivastava14a.html} {Dropout: A simple way to prevent neural networks from overfitting}.
\newblock \emph{Journal of Machine Learning Research}, 15(56):1929--1958.

\bibitem[{Treviso and Martins(2020)}]{treviso-martins-2020-explanation}
Marcos Treviso and Andr{\'e} F.~T. Martins. 2020.
\newblock \href {https://doi.org/10.18653/v1/2020.blackboxnlp-1.10} {The explanation game: Towards prediction explainability through sparse communication}.
\newblock In \emph{Proceedings of the Third BlackboxNLP Workshop on Analyzing and Interpreting Neural Networks for NLP}, pages 107--118, Online. Association for Computational Linguistics.

\bibitem[{Trockman et~al.(2024)Trockman, Harutyunyan, Kolter, Kumar, and Bhojanapalli}]{trockman2024mimeticinitializationhelpsstate}
Asher Trockman, Hrayr Harutyunyan, J.~Zico Kolter, Sanjiv Kumar, and Srinadh Bhojanapalli. 2024.
\newblock \href {https://arxiv.org/abs/2410.11135} {Mimetic initialization helps state space models learn to recall}.
\newblock \emph{Preprint}, arXiv:2410.11135.

\bibitem[{Vaswani et~al.(2017)Vaswani, Shazeer, Parmar, Uszkoreit, Jones, Gomez, Kaiser, and Polosukhin}]{vaswani2017attention}
Ashish Vaswani, Noam Shazeer, Niki Parmar, Jakob Uszkoreit, Llion Jones, Aidan~N Gomez, \L~ukasz Kaiser, and Illia Polosukhin. 2017.
\newblock \href {https://proceedings.neurips.cc/paper_files/paper/2017/file/3f5ee243547dee91fbd053c1c4a845aa-Paper.pdf} {Attention is all you need}.
\newblock In \emph{Advances in Neural Information Processing Systems}, volume~30. Curran Associates, Inc.

\bibitem[{Vilar et~al.(2006)Vilar, Popovic, and Ney}]{vilar-etal-2006-aer}
David Vilar, Maja Popovic, and Hermann Ney. 2006.
\newblock \href {https://aclanthology.org/2006.iwslt-papers.7} {{AER}: do we need to {``}improve{''} our alignments?}
\newblock In \emph{Proceedings of the Third International Workshop on Spoken Language Translation: Papers}, Kyoto, Japan.

\bibitem[{Vo et~al.(2025)Vo, Pham, Tong, and Nguyen}]{vo2025demystifying}
Thieu Vo, Duy-Tung Pham, Xin~T. Tong, and Tan~Minh Nguyen. 2025.
\newblock \href {https://openreview.net/forum?id=qtTIP5Gjc5} {Demystifying the token dynamics of deep selective state space models}.
\newblock In \emph{The Thirteenth International Conference on Learning Representations}.

\bibitem[{Wiegreffe and Pinter(2019)}]{wiegreffe-pinter-2019-attention}
Sarah Wiegreffe and Yuval Pinter. 2019.
\newblock \href {https://doi.org/10.18653/v1/D19-1002} {Attention is not not explanation}.
\newblock In \emph{Proceedings of the 2019 Conference on Empirical Methods in Natural Language Processing and the 9th International Joint Conference on Natural Language Processing (EMNLP-IJCNLP)}, pages 11--20, Hong Kong, China. Association for Computational Linguistics.

\bibitem[{Wu and He(2018)}]{Wu_2018_ECCV}
Yuxin Wu and Kaiming He. 2018.
\newblock Group normalization.
\newblock In \emph{Proceedings of the European Conference on Computer Vision (ECCV)}.

\bibitem[{Xu et~al.(2024)Xu, Yang, Wang, Du, and Chen}]{xu2024survey}
Rui Xu, Shu Yang, Yihui Wang, Bo~Du, and Hao Chen. 2024.
\newblock \href {https://arxiv.org/abs/2404.18861v1} {A survey on vision mamba: Models, applications and challenges}.
\newblock \emph{arXiv preprint arXiv:2404.18861v1}.

\bibitem[{Yang et~al.(2024)Yang, Wang, Zhang, Shen, and Kim}]{yang2024parallelizing}
Songlin Yang, Bailin Wang, Yu~Zhang, Yikang Shen, and Yoon Kim. 2024.
\newblock \href {https://openreview.net/forum?id=y8Rm4VNRPH} {Parallelizing linear transformers with the delta rule over sequence length}.
\newblock In \emph{The Thirty-eighth Annual Conference on Neural Information Processing Systems}.

\end{thebibliography}

\appendix

\clearpage

\section{Hidden Attention Derivation in Mamba}
\label{sec:derivation_m}

This appendix provides a detailed derivation of the hidden‐attention matrix $\bm{M}$ in both Mamba-1 and Mamba-2, showing how their element-wise recurrences can be written in the form $\bm{\Upsilon} = \bm{M}\,\bm{X}$.

\subsection{Mamba-1 Derivation}
\label{sec:derivation_m_mamba1}

Recall the Mamba-1 recurrence (ignoring skip connections) for each time step $i\ge1$:
\begin{align*}
    \bm{H}_i &= \bm{A}_i \odot \bm{H}_{i-1} + \bm{B}_i \odot \bm{X}_i &\in \mathbb{R}^{R \times D}, \\
    \bm{\upsilon}_i &= \bm{H}_i^\top \bm{c}_i &\in \mathbb{R}^{D},
\end{align*}
where $\bm{X}_i = \bm{1}_r\bm{x}_i^\top \in \mathbb{R}^{R \times D}$ is an $R$-sized stack of the input, $\bm{A}_i \in \mathbb{R}^{R \times D}$ represents $D$ diagonal matrices of size $R \times R$, $\bm{B}_i \in \mathbb{R}^{R \times D}$, $\bm{c}_i \in \mathbb{R}^{R}$, and $\odot$ is the Hadamard product.
Setting $\bm{H}_0 = \bm{0}$, we can unroll the recurrence to see how past tokens contribute:
\begin{align*}
    \bm{H}_1 &= \bm{A}_1 \odot \bm{0} + \bm{B}_1 \odot \bm{X}_1.
    \\ \nonumber \\
    \bm{H}_2 &= \bm{A}_2 \odot \bm{H}_1 + \bm{B}_2 \odot \bm{X}_2 \\
             &= \bm{A}_2 \odot (\bm{A}_1 \odot \bm{0} + \bm{B}_1 \odot \bm{X}_1) + \bm{B}_2 \odot \bm{X}_2 \\
             &= \bm{A}_2 \odot \bm{B}_1 \odot \bm{X}_1 + \bm{B}_2 \odot \bm{X}_2.
    \\ \nonumber \\
    \bm{H}_3 &= \bm{A}_3 \odot \bm{H}_2 + \bm{B}_3 \odot \bm{X}_3 \\
             &= \bm{A}_3 \odot \left[ \bm{A}_2 \odot \bm{B}_1 \odot \bm{X}_1 + \bm{B}_2 \odot \bm{X}_2 \right] \\ &\quad + \bm{B}_3 \odot \bm{X}_3 \\
             &= \bm{A}_3 \odot \bm{A}_2 \odot \bm{B}_1 \odot \bm{X}_1 \\ &\quad + \bm{A}_3 \odot \bm{B}_2 \odot \bm{X}_2 + \bm{B}_3 \odot \bm{X}_3.
\end{align*}
Hence, in general for any $i$, we have:
\begin{align*}
    \bm{H}_i &= \sum_{j=1}^i \left(\oprod_{k=j+1}^i \bm{A}_k \right) \odot \bm{B}_j \odot \bm{X}_j  &\in \mathbb{R}^{R \times D}, \\
    \bm{\upsilon}_i &= \bm{H}_i ^ \top \bm{c}_i  &\in \mathbb{R}^{D},
\end{align*}
where we write $\oprod$ to indicate an element-wise product over the indices $k$. 

\paragraph{Block‐matrix expression.}
To capture this in matrix form, observe that each coordinate of $\bm{X}_j$ gets multiplied by a chain of element-wise factors $\bm{A}_k$ and $\bm{B}_j$, then finally projected by $\bm{c}_i$.  Aggregating these dimension-wise scalars into a diagonal matrix $\bm{M}_{i,j}\in\mathbb{R}^{D\times D}$ yields
\begin{align*}
    \bm{M}_{i,j} = 
        \text{Diag} \left( \left[ \left(\oprod\limits_{k=j+1}^{i} \bm{A}_k \right) \odot \bm{B}_j \right]^{\top} \bm{c}_i \right),
\end{align*}
for all $j \leq i$, and $\bm{M}_{i,j} = \bm{0}$ otherwise.
Stacking these $\bm{M}_{i,j}$ blocks into a 4D tensor $\bm{M}\in\mathbb{R}^{N\times N\times D\times D}$ gives us
\begin{equation}
  \bm{\Upsilon} \;=\; \bm{M}\,\bm{X}, 
\end{equation}
once we interpret $\bm{M}$ as an $N\times N$ grid of $D\times D$ blocks and flatten $\bm{X}\in\mathbb{R}^{N\times D}$ to a length‐$(ND)$ vector, as explained below in \S\ref{sec:block_matrix_form}.

\subsection{Mamba-2 Derivation}
\label{sec:derivation_m_mamba2}

Mamba-2 uses a similar idea but modifies $\bm{A}_i$ into a \emph{diagonal matrix} of size $R\times R$, rather than an element-wise parameter array.  Formally,
\begin{align*}
    \bm{H}_i &=\bm{A}_i \bm{H}_{i-1} + \bm{B}_i \odot \bm{X}_i &\in \mathbb{R}^{R \times D}, \\
    \bm{\upsilon}_i &= \bm{H}_i^\top\bm{c}_i &\in \mathbb{R}^{D},
\end{align*}
where $\bm{A}_i=a_i\,\bm{I}_{R\times R}$. Unrolling similarly, we get
\begin{align*}
    \bm{H}_t &= \sum_{j=1}^t \left(\prod_{k=j+1}^t \bm{A}_k \right) \odot \bm{B}_j \odot \bm{X}_j  &\in \mathbb{R}^{R \times D}, \\
    \bm{\upsilon}_t &= \bm{H}_t ^ \top \bm{c}_t  &\in \mathbb{R}^{D}.
\end{align*}
Since each $\bm{A}_k$ is a diagonal matrix, the product $\prod_{k=j+1}^i \bm{A}_k$ remains diagonal.  Aggregating the resulting dimension-wise multipliers again forms $\bm{M}_{i,j}\in\mathbb{R}^{D\times D}$, leading to
\begin{align*}
    \bm{M}_{i,j} =
        \text{Diag} \left( \left[ \left(\prod\limits_{k=j+1}^{i} \bm{A}_k \right) \odot \bm{B}_j \right]^{\top} \bm{c}_i \right),
\end{align*}
for all $j \leq i$, and $\bm{M}_{i,j} = \bm{0}$ otherwise. 
The shape‐flattening for $\bm{M}$ and $\bm{X}$ then follows the same block‐matrix logic as in Mamba-1.

\subsection{Block‐Matrix Implementation}
\label{sec:block_matrix_form}

Define the overall 4D tensor $\bm{M}\in\mathbb{R}^{N\times N\times D\times D}$ by gathering the blocks $\bm{M}_{i,j}$ from above.  In matrix form, we can treat $\bm{M}$ as an $N\times N$ grid of $D\times D$ blocks, thus flattening to $\bm{M}\in\mathbb{R}^{(ND)\times(ND)}$.  Simultaneously, reshape $\bm{X}\in\mathbb{R}^{N\times D}$ into a vector of length $(ND)$ by stacking each token row.  Then the usual matrix–vector product recovers the unrolled recurrence:
\begin{equation*}
    \bm{\Upsilon} = \bm{M} \bm{X} \Leftrightarrow \bm{\upsilon}_i = \sum_{j=1}^i \bm{M}_{i,j}\,\bm{x}_j.
\end{equation*}
Concretely, the $i$\textsuperscript{th} block row of $\bm{M}$ multiplies the token embeddings $\{\bm{x}_j\}_{j=1}^N$, and the result is then reshaped back to produce an $N\times D$ matrix, whose $i$\textsuperscript{th} row is precisely $\bm{\upsilon}_i^\top$.

\begin{figure*}[t]
  \includegraphics[width=0.495\textwidth]{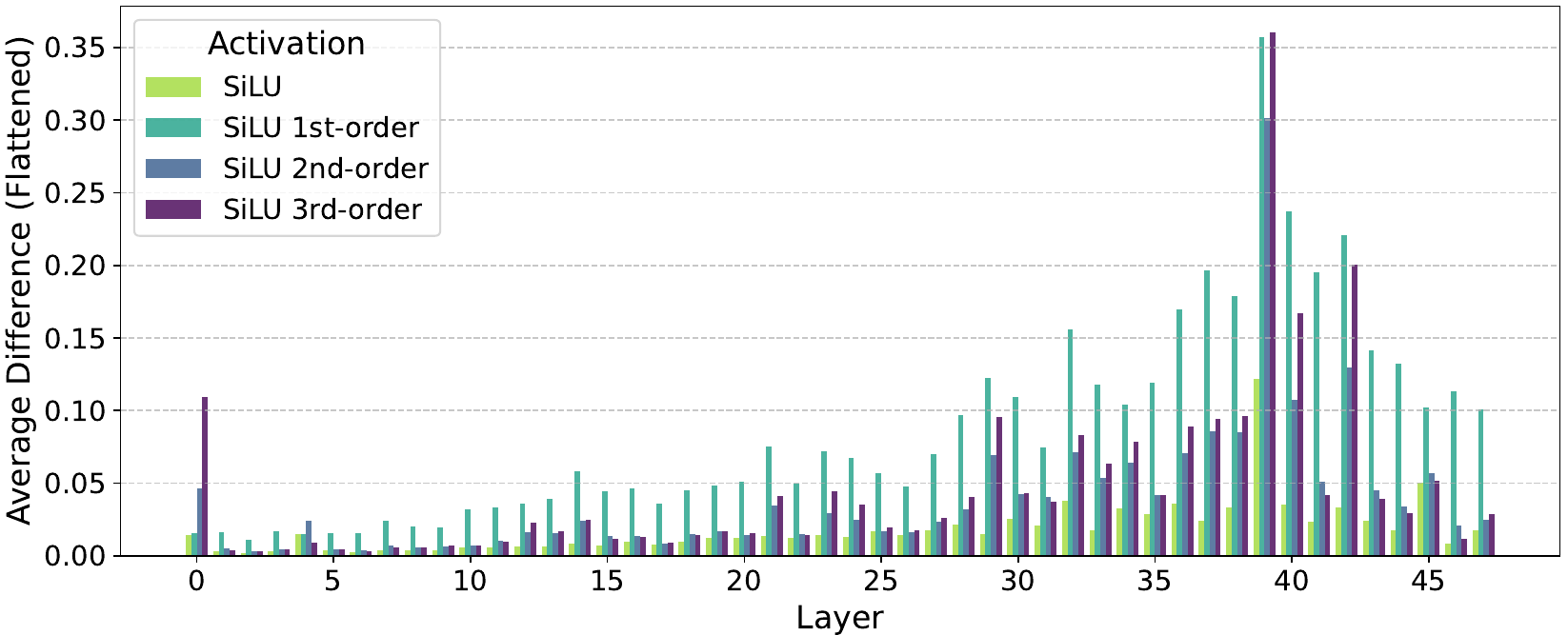}
  \includegraphics[width=0.49\textwidth]{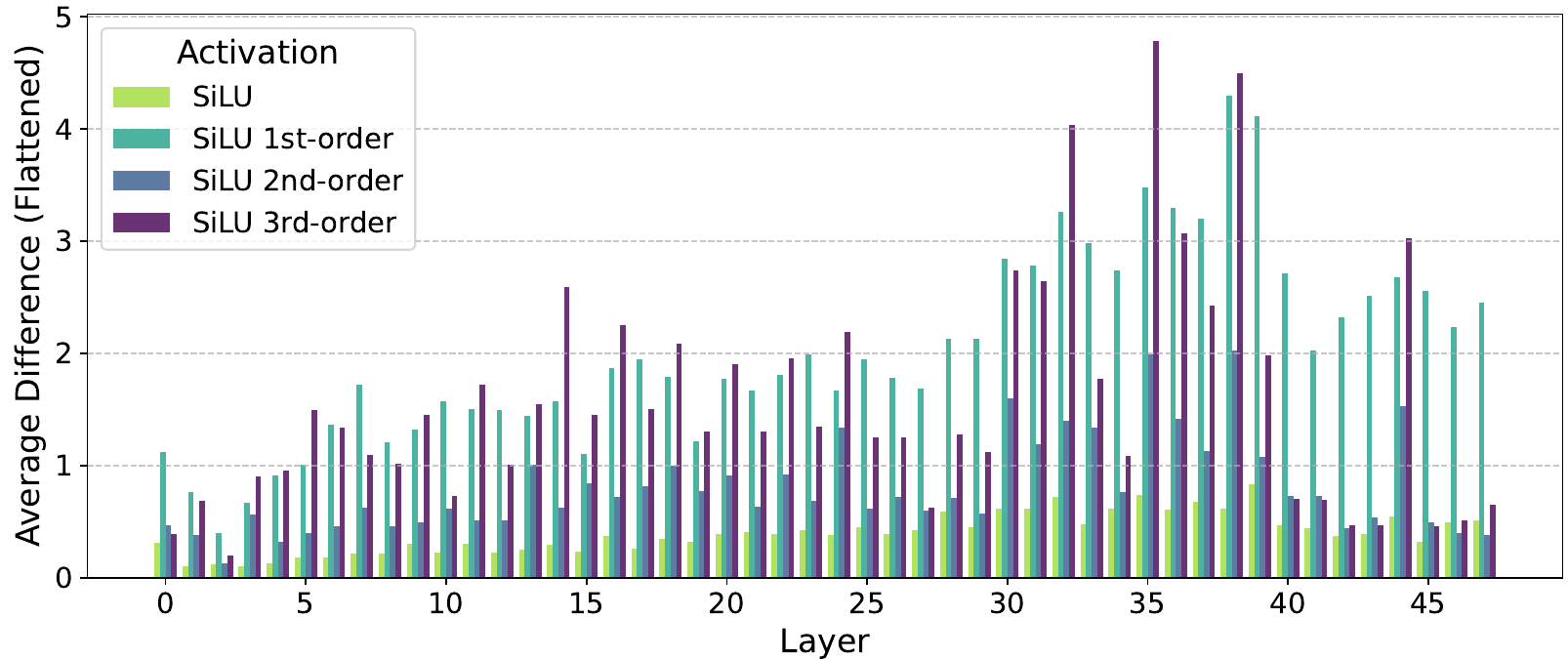}
  \caption{Error amounting to the average difference between the regular Mamba-1 (left) and Mamba-2 (right) layer output and the interpretable version with different approximations $f$ in Equation~\ref{eq:f_additive_decomp_mamba}.}
  \label{fig:mamba-mt-act-approx-error}
\end{figure*}

\section{Experimental Details}
\label{sec:exp-details}

\subsection{Copying}
\label{sec:app_copying}
We use 8-layer Mamba 1 and 2 models with 512 as the hidden size and 32 as the vocabulary size, the state dimension is set to 16 and 128 for Mamba 1 and 2, respectively. Only layer 4 is initialized as per \citet{trockman2024mimeticinitializationhelpsstate}, with their optimal configuration (which differs from Mamba 1 to 2). Optimization: AdamW \citep{loshchilov2019decoupled} optimizer with the inverse square root \citep{vaswani2017attention} learning rate scheduler (500 warmup steps, 5000 total steps, 256 samples per batch) and a learning rate of $7e-4$. No dropout or gradient clipping was used.
The copying dataset was generated as per \citep{jelassi2024repeat} and contains 5000 training samples and 128 evaluation samples.

\subsection{Machine Translation} 
\label{sec:app_machine_translation}
All model dimensions are coupled to their officially released checkpoints. Optimization:  AdamW \citep{loshchilov2019decoupled} optimizer with a cosine learning rate scheduler (2000 warmup steps, 18000 steps, 64 samples per batch) and a learning rate of $7e-4$. Dropout \citep{JMLR:v15:srivastava14a} rate was set to 0.3 and no gradient clipping was used.
The IWSLT17 \citep{cettolo-etal-2017-overview} dataset contains 232825 training samples, 890 validation samples and 8597
samples for both the \textsc{de$\leftrightarrow$en} and \textsc{fr$\leftrightarrow$en} versions.

\subsection{Approximation Error}
\label{sec:app_aproximation_error}
All model dimensions are coupled to their officially released checkpoints when performing continued language pretraining. Optimization: AdamW \citep{loshchilov2019decoupled} optimizer with a WSD \citep{hu2024minicpmunveilingpotentialsmall} learning rate scheduler (2000 warmup steps, 27900 stable steps, 3100 decay steps, 32k tokens per batch) and a learning rate of $5e-5$. We used gradient clipping set to $5.0$ and no dropout. 
Moreover, we employed an $\alpha$ parameter in order to smoothly interpolate between the old (SiLU) and the new activations (ReLU or identity).
The value of $\alpha$ followed a power law during training: $\textsf{min}(1, \textsf{current\_step} / (\textsf{total\_steps} - \textsf{decay\_steps}))^2$. 
Note that the learning rate decay period coincides with the phase where the model relies only on the new activation.

\subsection{Computational Details}

All experiments involving \methodname were carried on Nvidia RTX A6000 GPUs with 48GB VRAM.

\section{Extended Approximation Error}
\label{sec:extended-error}

Following \S\ref{subsec:approximation_error_analysis}, we include additional data which details how the decomposition error changes with different \textsf{SiLU} approximations $f$ on each layer. 
This experiment has been conducted over the GoldAlign \citep{vilar-etal-2006-aer} dataset. The results can be seen in Figure~\ref{fig:mamba-mt-act-approx-error}. Overall, casting $f$ as \textsf{SiLU} leads to the lowest approximation errors across all models and layers.

\section{Additional Experiments}
\label{sec:additional-experiments}

\subsection{Copying}
\label{subsec:app-copying}

\begin{figure*}[t]
  \centering
    \includegraphics[width=0.195\textwidth]{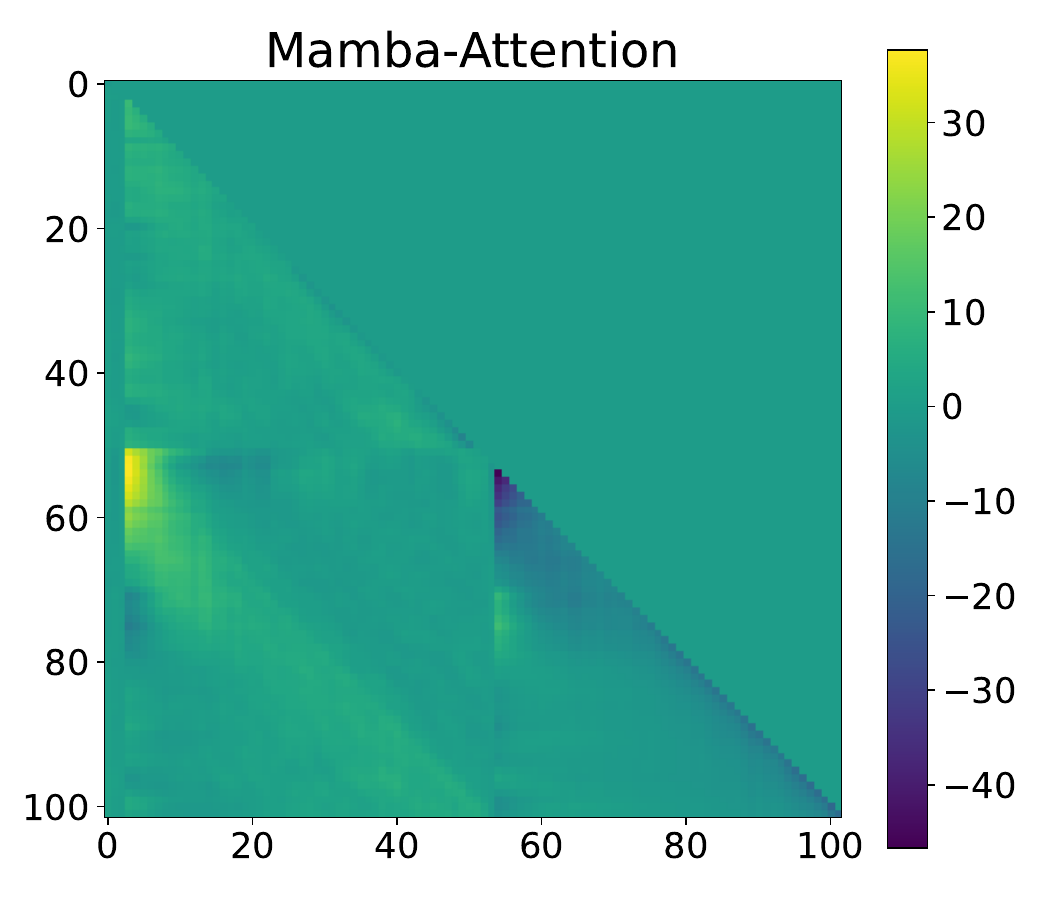}
    \hfill
    \includegraphics[width=0.195\textwidth]{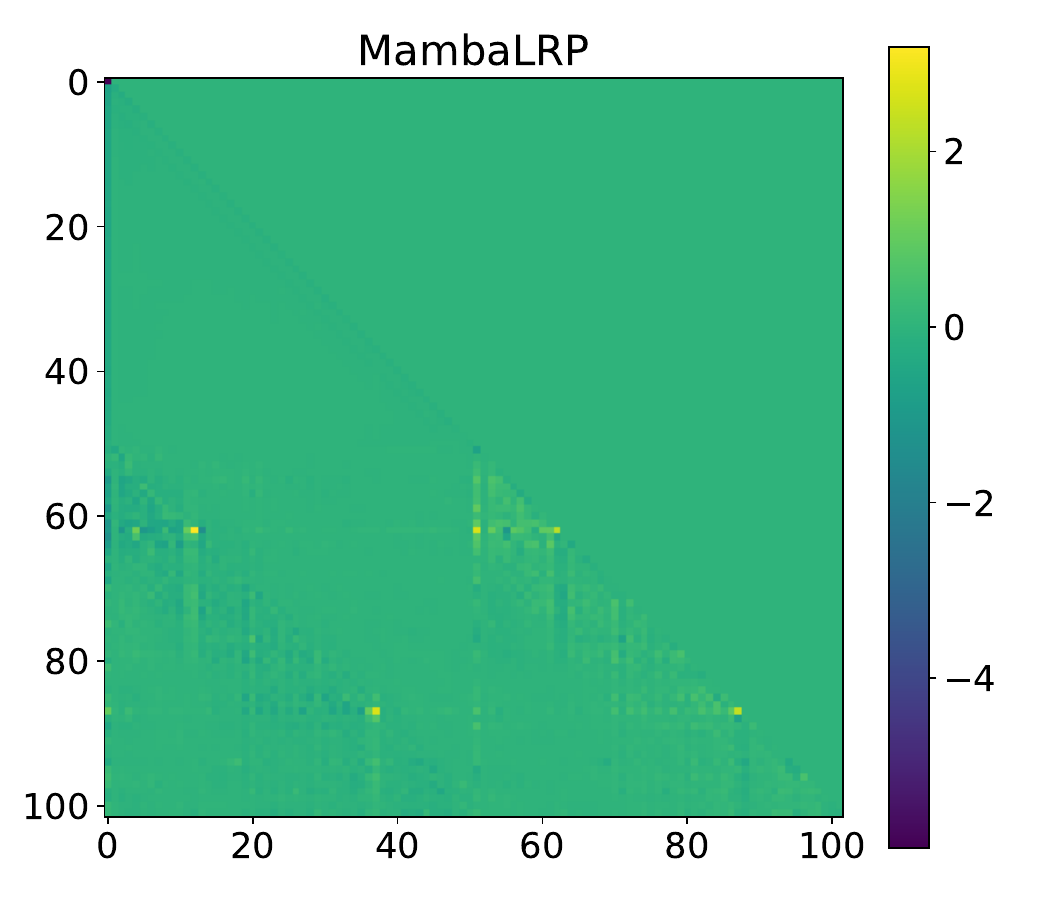}
    \hfill
    \includegraphics[width=0.195\textwidth]{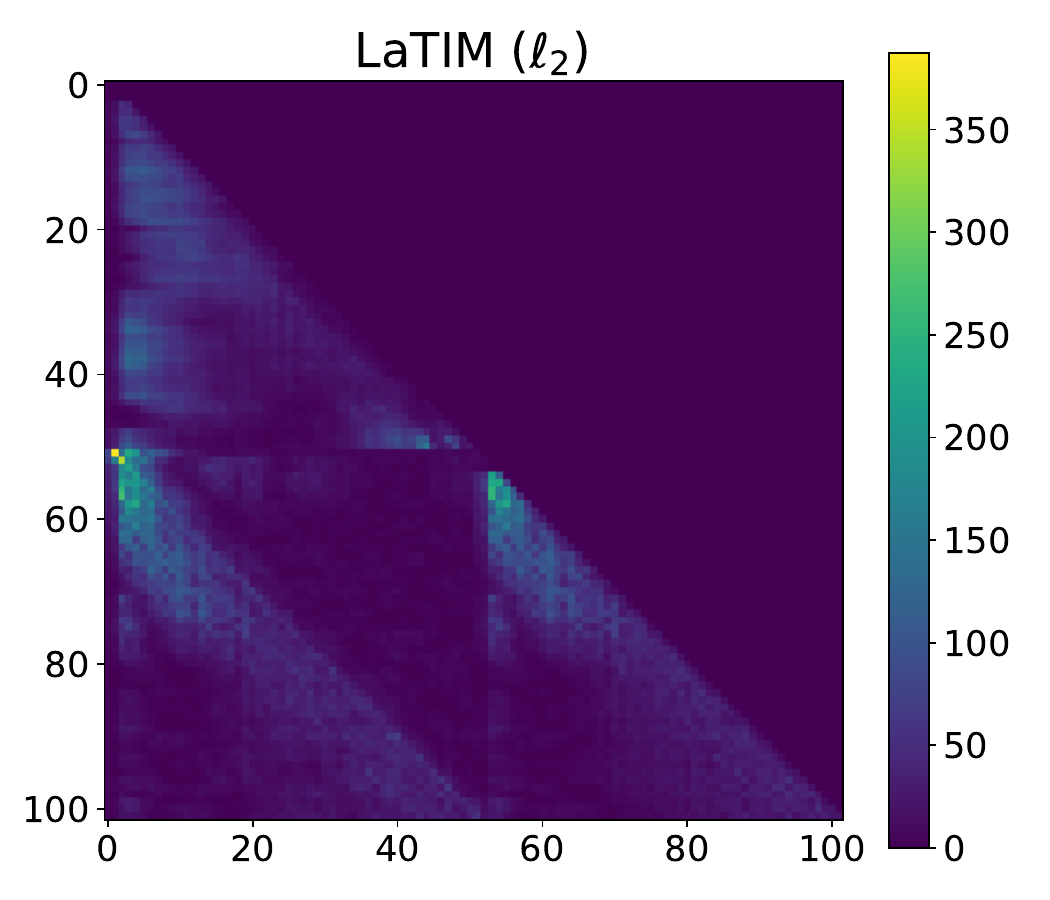}
    \hfill
    \includegraphics[width=0.195\textwidth]{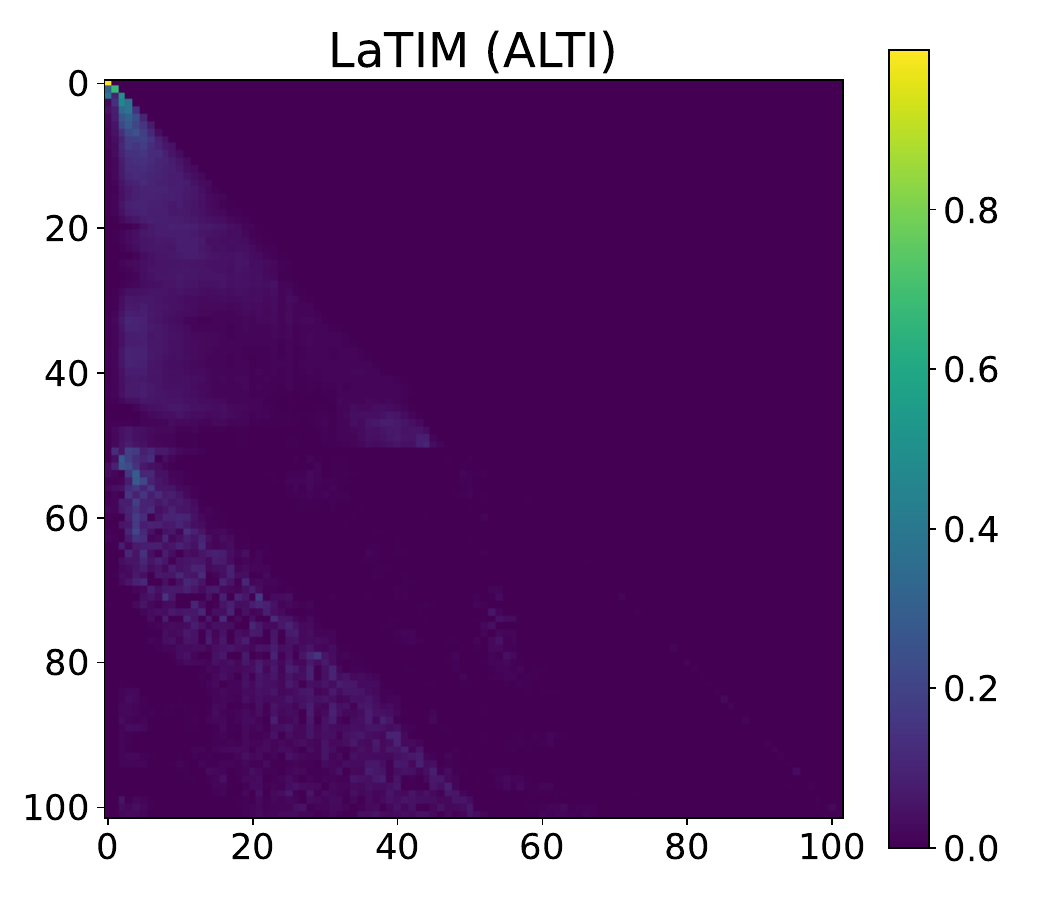}
    \hfill
    \includegraphics[width=0.195\textwidth]{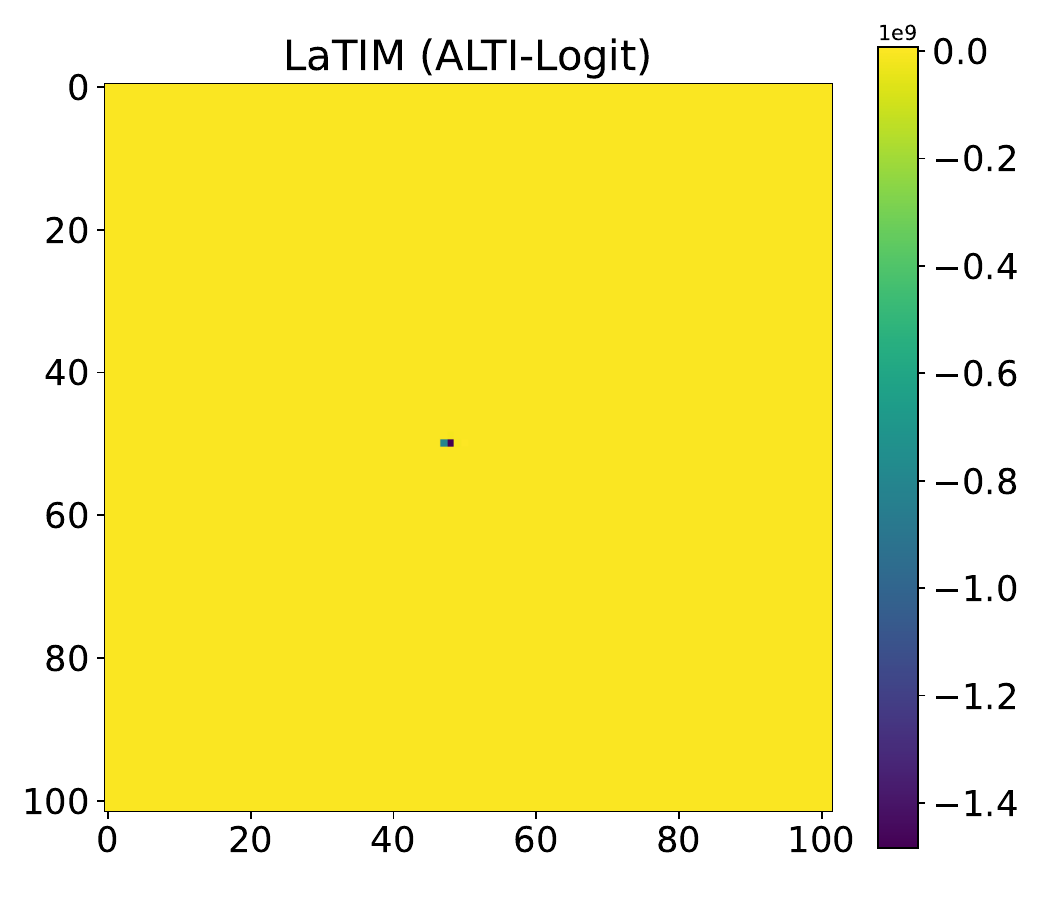}
 \caption{Heatmaps produced by different interpretable approaches for Mamba-1. 
 The interaction between source and copied tokens (along the diagonal line) becomes clearer with \methodname. 
 }
  \label{fig:mamba-copy-alti-vs-avg}
\end{figure*}

\begin{figure*}[t]
    \includegraphics[width=0.195\textwidth]{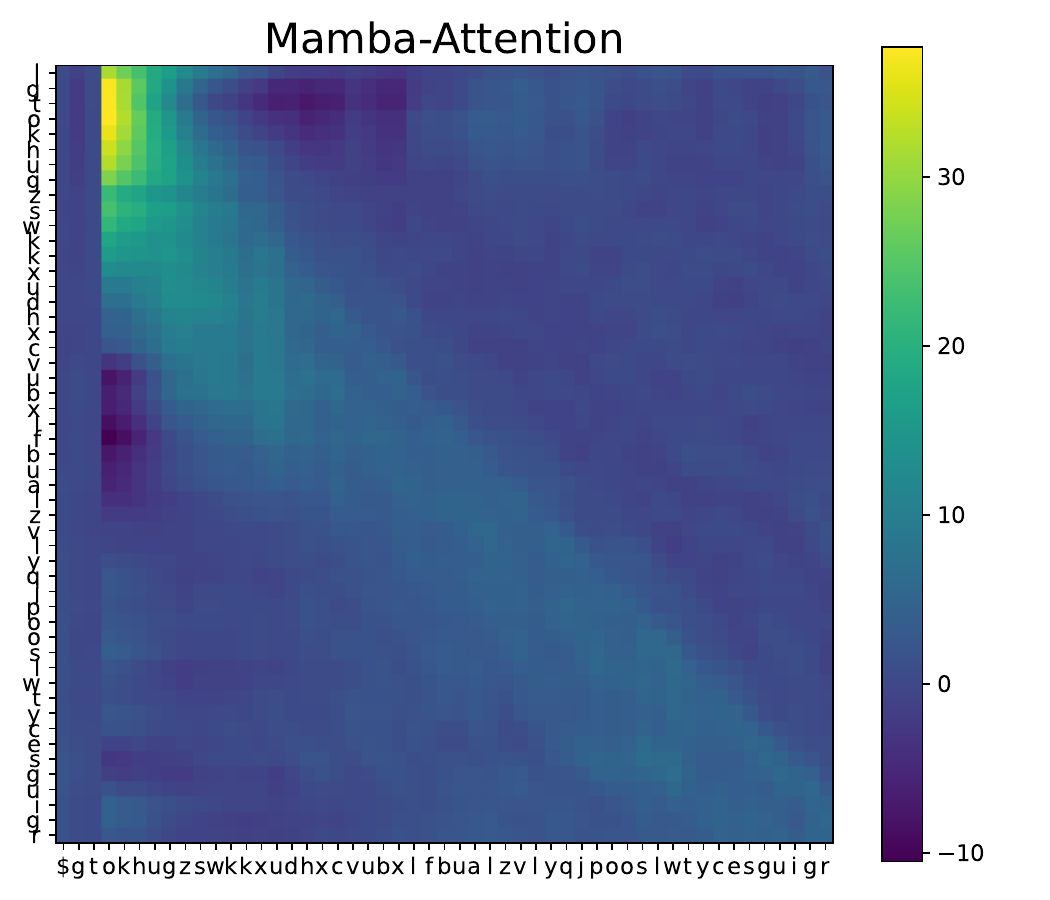}
    \hfill
    \includegraphics[width=0.195\textwidth]{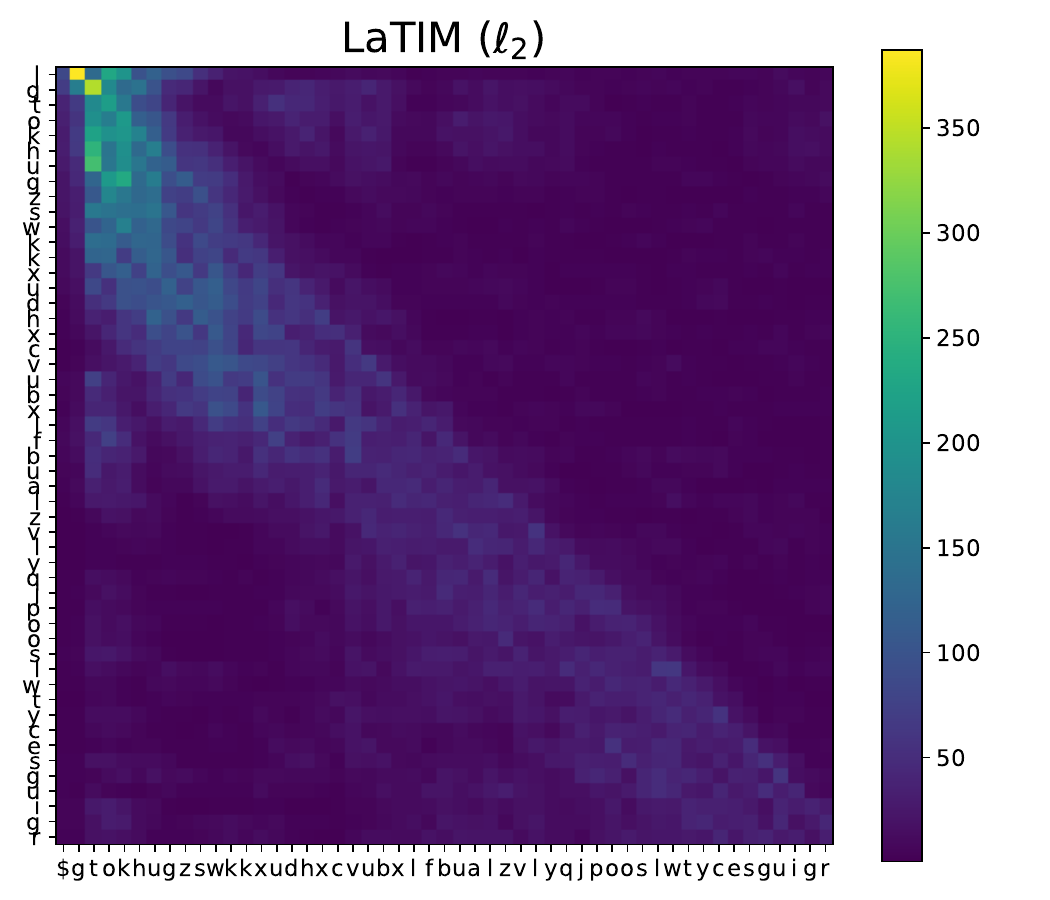}
    \hfill
    \includegraphics[width=0.195\textwidth]{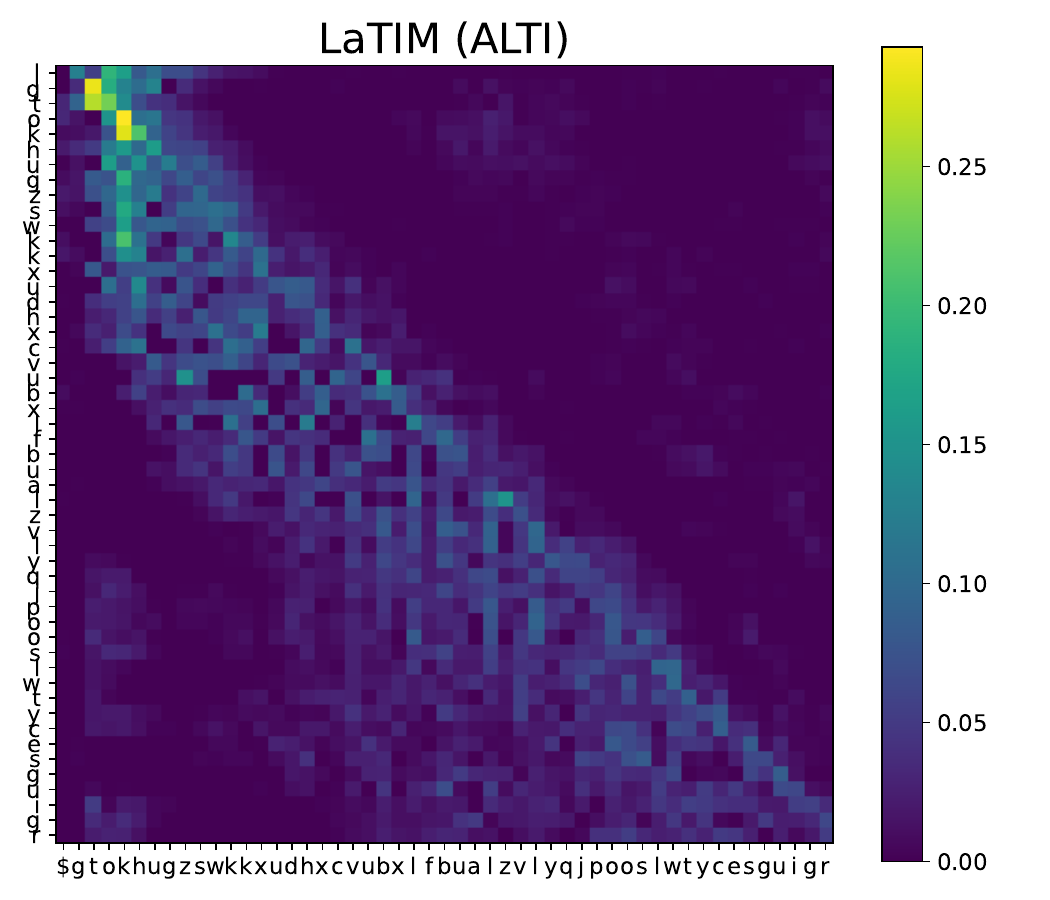}
    \hfill
    \includegraphics[width=0.195\textwidth]{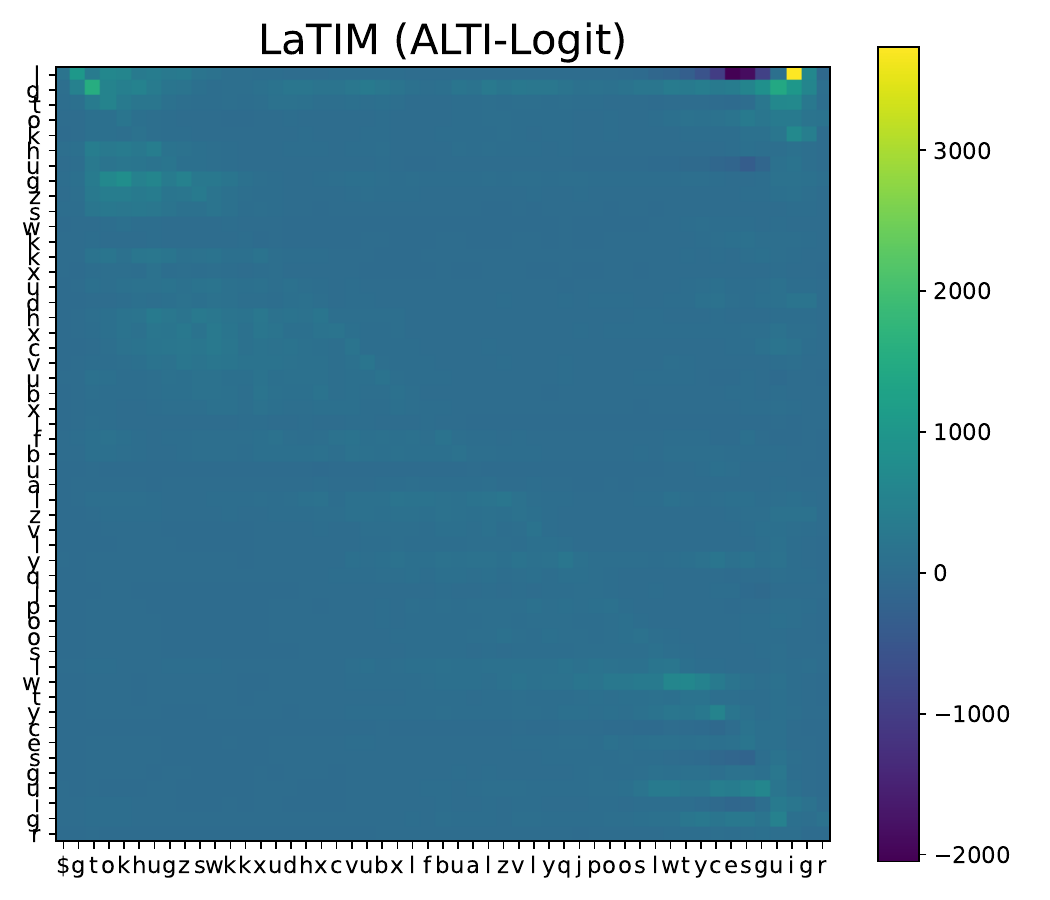}
    \centering
    \includegraphics[width=0.195\textwidth]{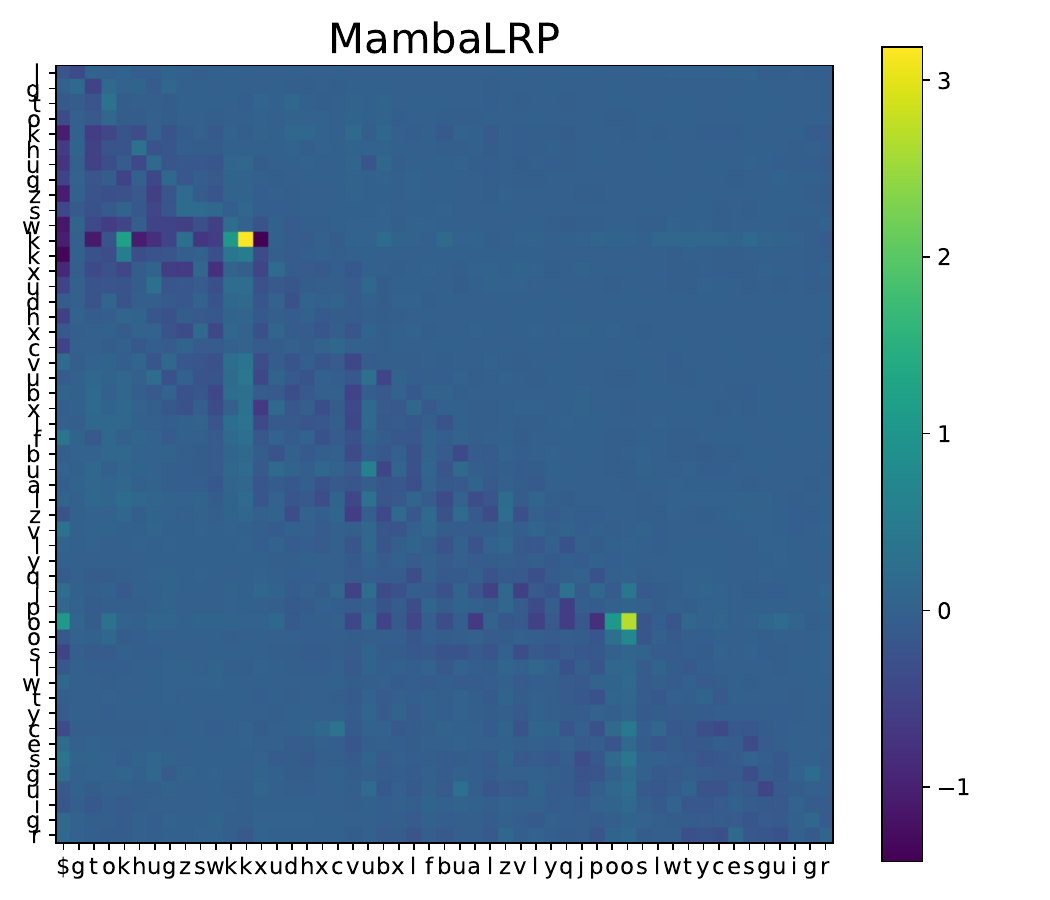}
    \\
    \includegraphics[width=0.195\textwidth]{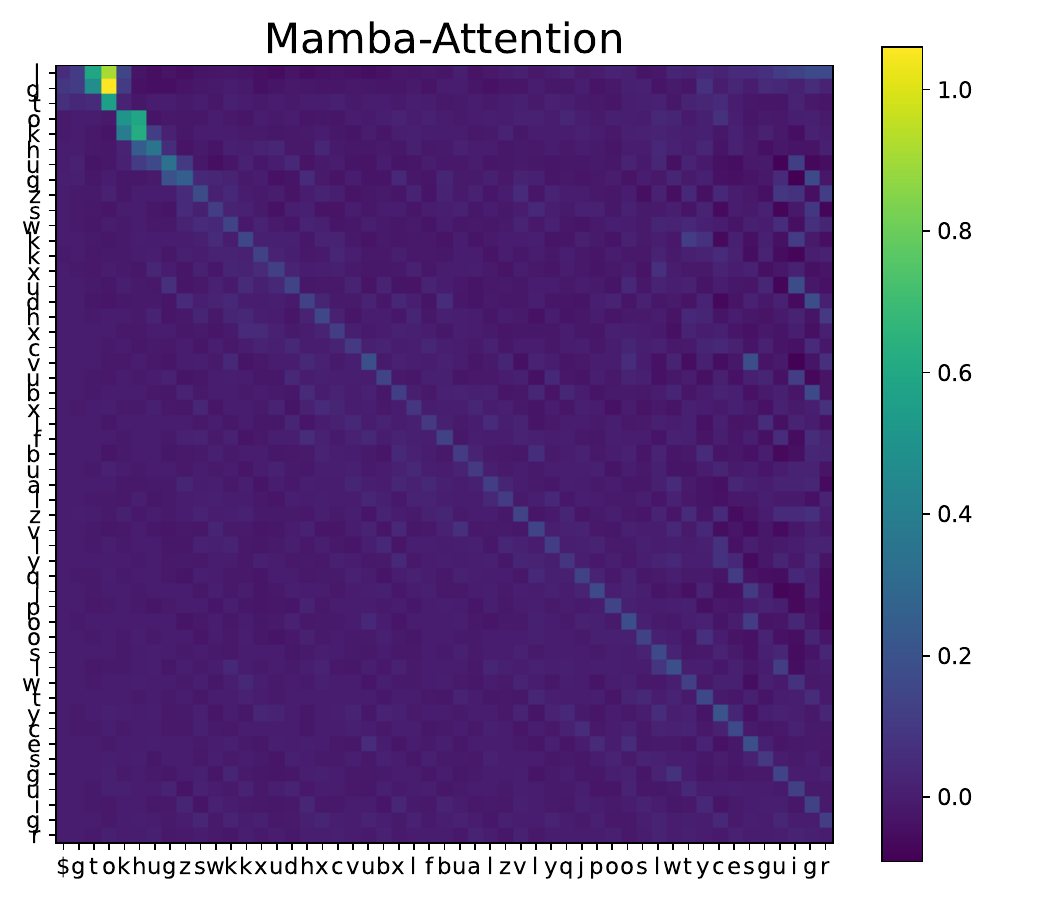}
    \ 
    \includegraphics[width=0.195\textwidth]{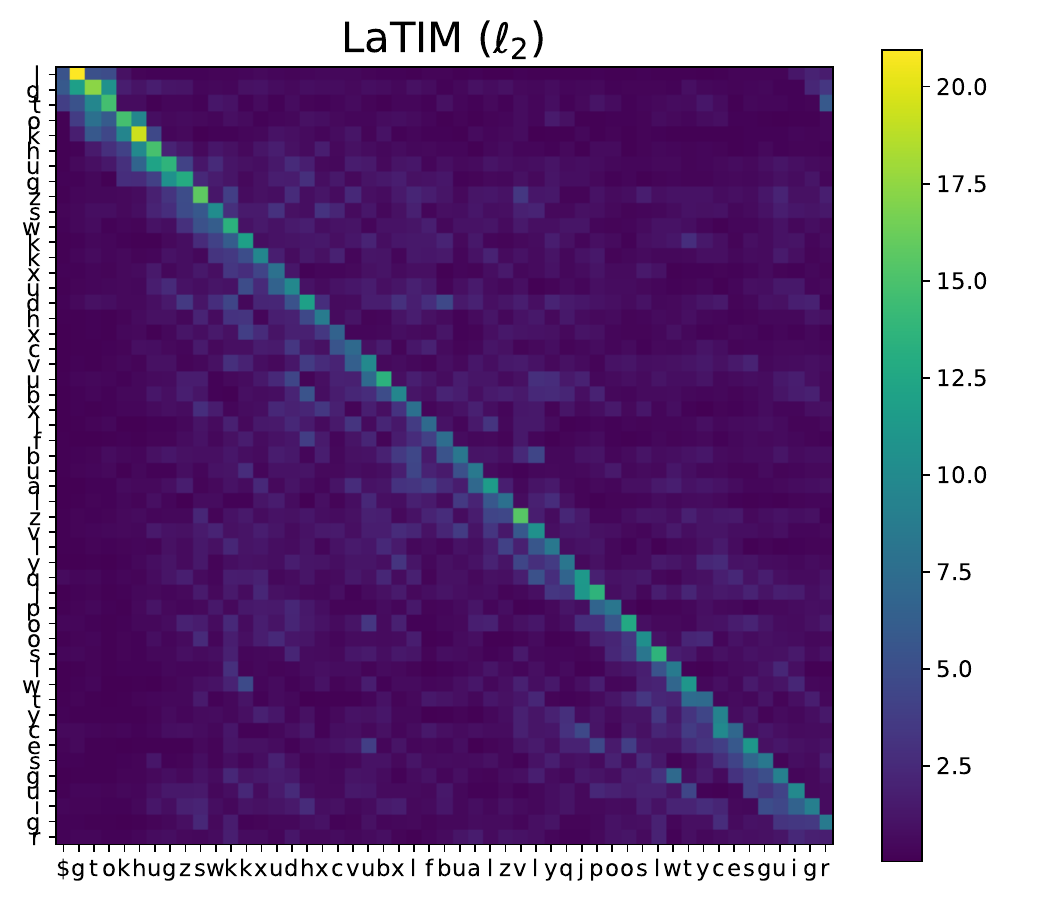}
    \ 
    \includegraphics[width=0.195\textwidth]{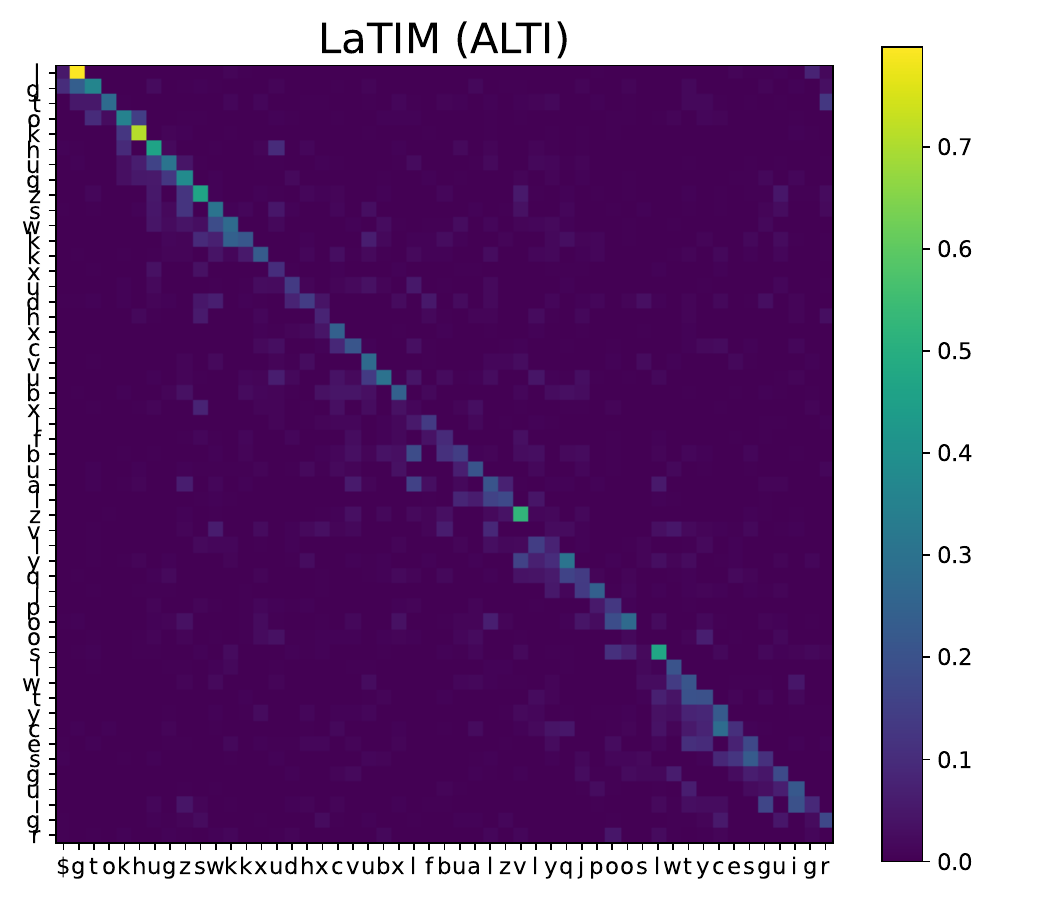}
    \ 
    \includegraphics[width=0.195\textwidth]{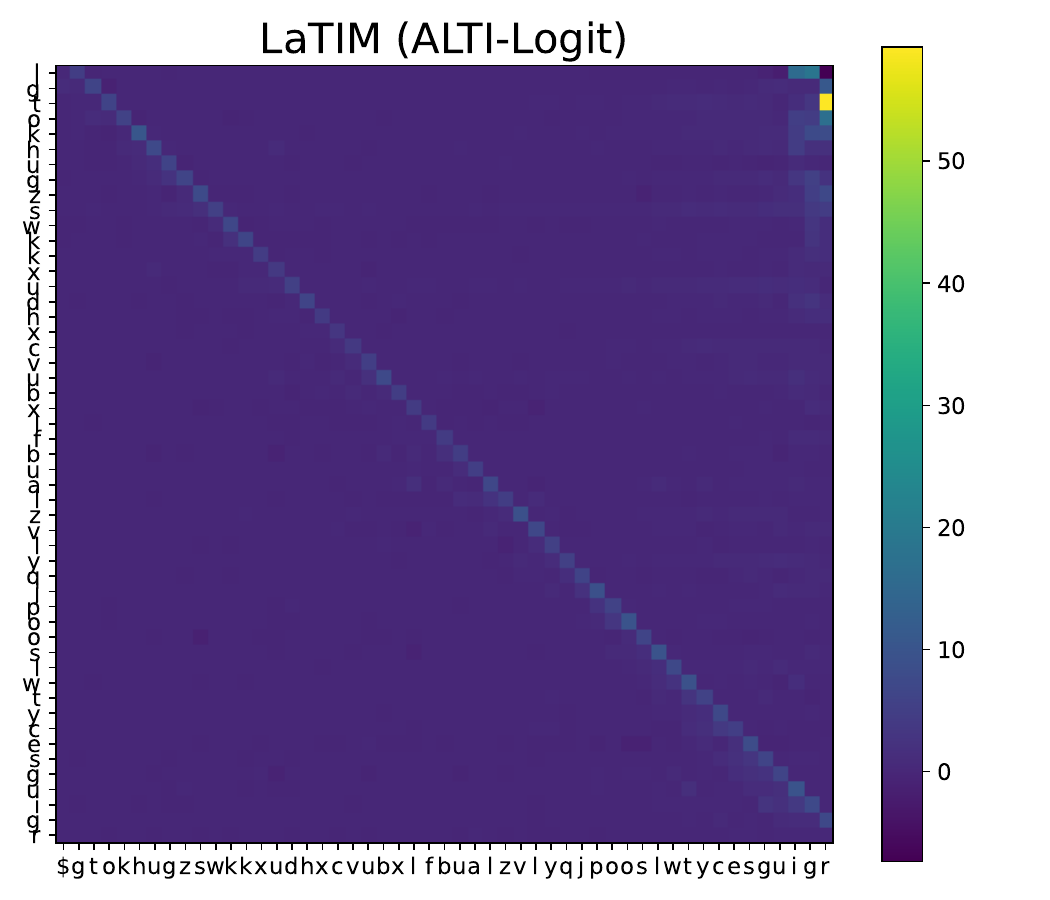}

  \caption{Heatmaps produced by the different interpretability methods for Mamba-1 (top) and Mamba-2 (bottom) on a copying sample after filtering the source$\rightarrow$copy interaction. Note how both models learned to focus over a off-diagonal pattern instead of a direct token-copy map.
  }
  \label{fig:filtered-copying}
\end{figure*}

In addition to the Mamba-2 visualizations in \S\ref{sec:copying}, we further include Mamba-1-based versions in Figure~\ref{fig:mamba-copy-alti-vs-avg}. These include a comparison with MambaLRP which performs especially poorly for this experiment as previously observed in Table~\ref{tab:copy_faithfulness}. 
Moreover, in Figure~\ref{fig:filtered-copying}  we show a filtered version of these plots with just the source$\rightarrow$copy interactions (left-bottom block). We highlight how models learn a pattern centered around the diagonal. 
As per this argument, Table~\ref{tab:copy_faithfulness} relies on the three main diagonals as its gold label.

\subsection{Machine Translation}
 \label{subsec:app-mt}

\begin{figure*}[t]
    \includegraphics[width=0.195\textwidth]{figs/mt/Mamba-Attention.pdf}
    \hfill
    \includegraphics[width=0.195\textwidth]{figs/mt/lrp.pdf}
    \hfill
    \includegraphics[width=0.195\textwidth]{figs/mt/LaTIM-l2.pdf}
    \hfill
    \includegraphics[width=0.195\textwidth]{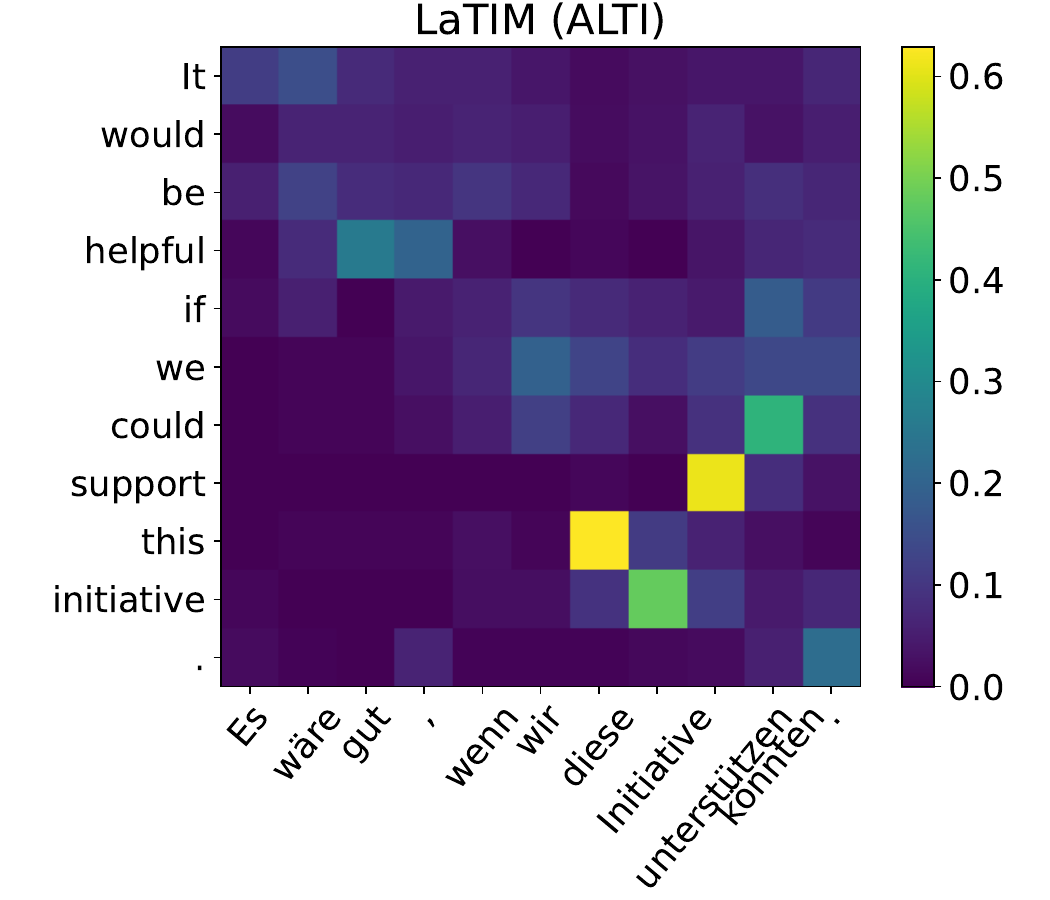}
    \includegraphics[width=0.195\textwidth]{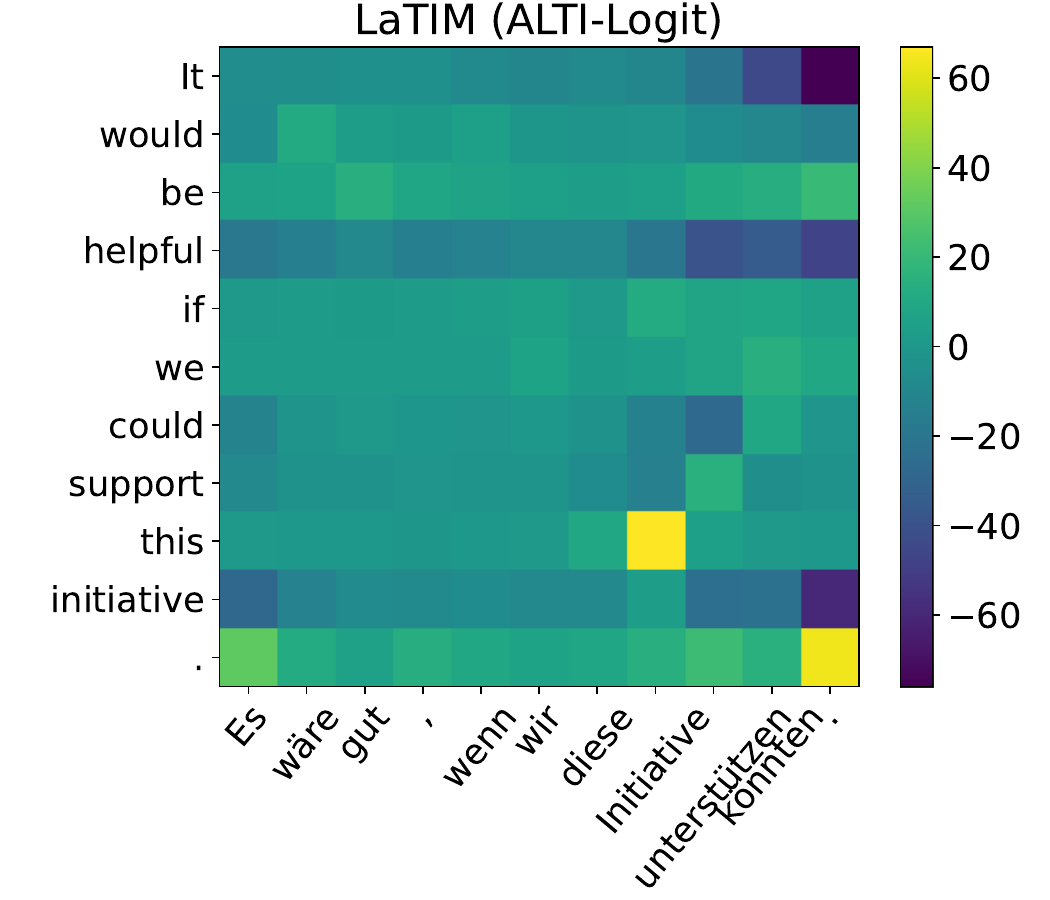}
    \\
    \includegraphics[width=0.195\textwidth]{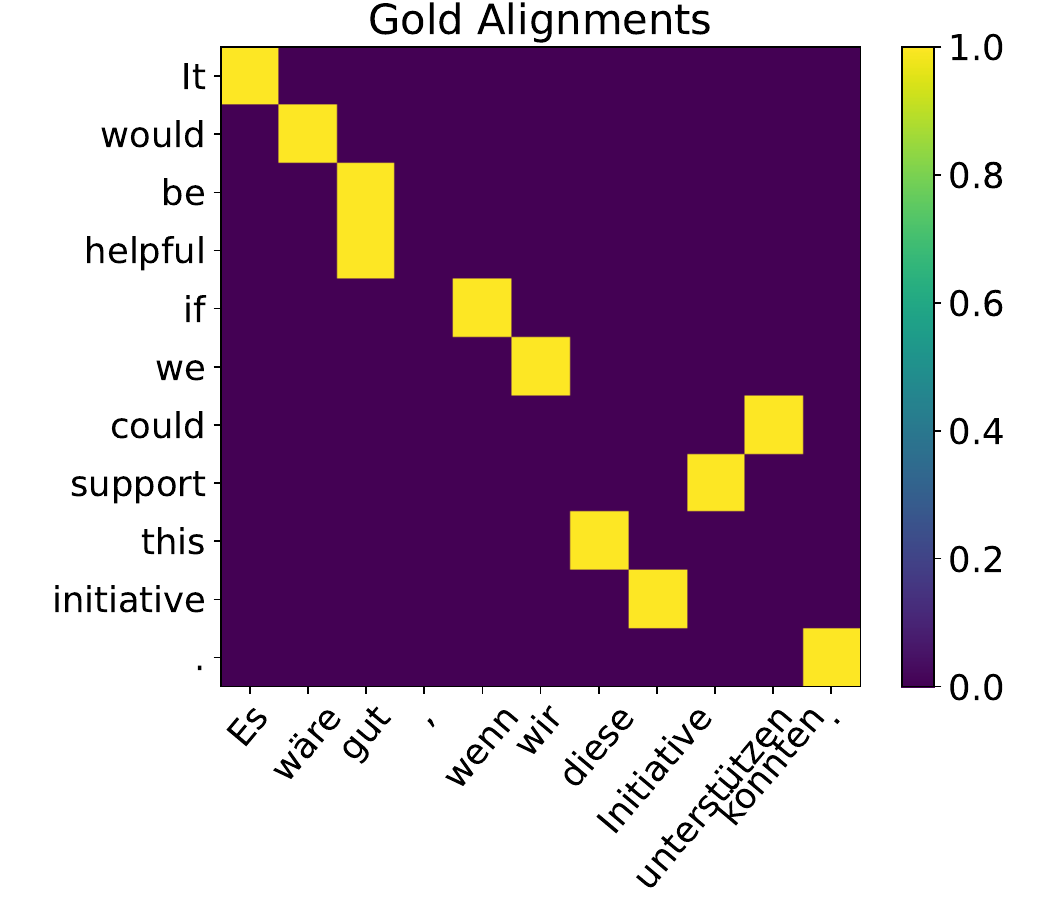}  
    \hfill
    \includegraphics[width=0.195\textwidth]{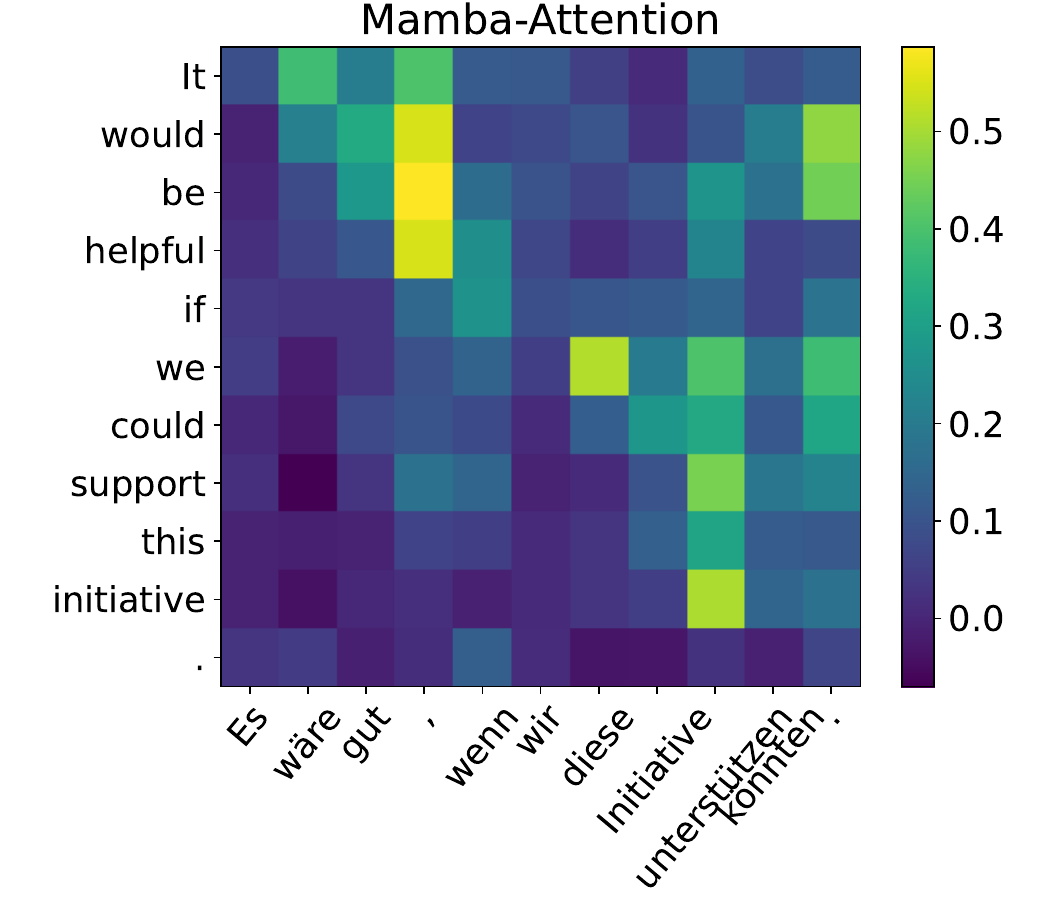}
    \hfill
    \includegraphics[width=0.195\textwidth]{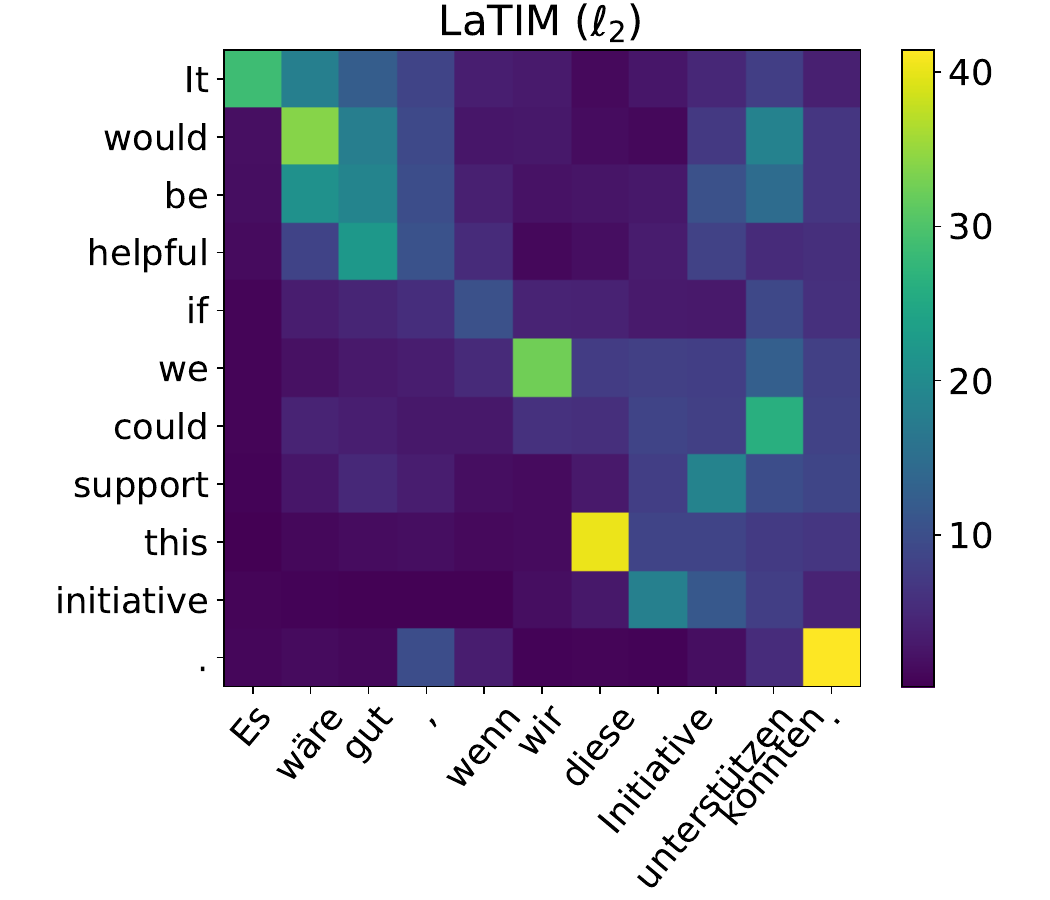}
    \hfill
    \includegraphics[width=0.195\textwidth]{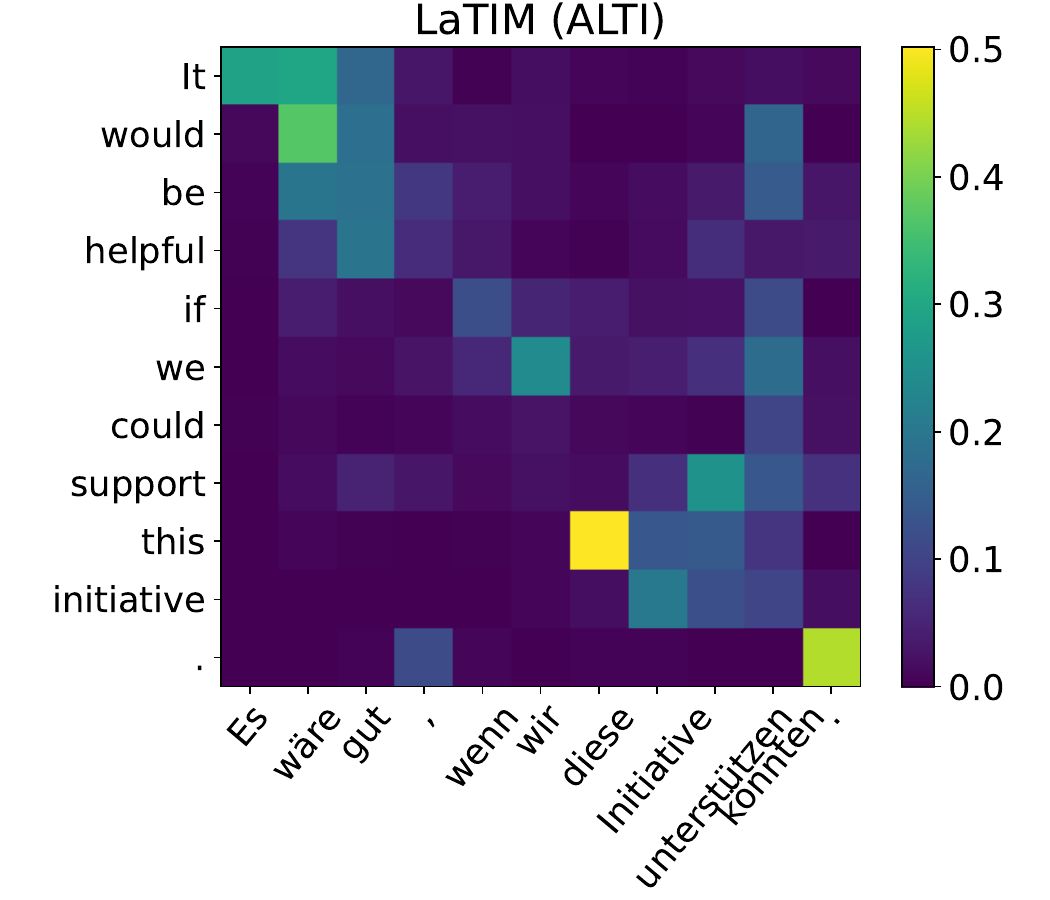}
    \hfill
    \includegraphics[width=0.195\textwidth]{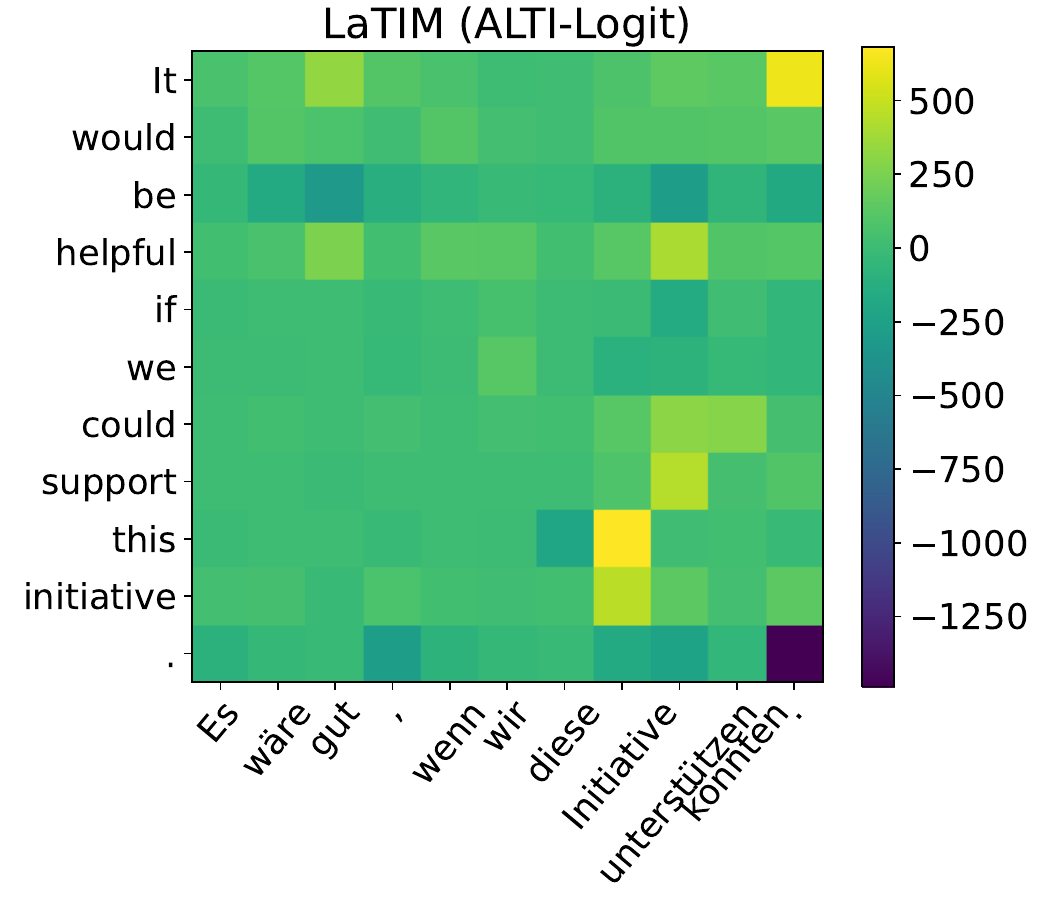}
  \caption{Heatmaps produced by the different interpretability methods for Mamba-1 (top) and Mamba-2 (bottom) fine-tuned on \textsc{de$\rightarrow$en} data. 
  }
  \label{fig:mamba2-mt-alti-vs-avg}
\end{figure*}

In addition to the Mamba-1 visualizations in \S\ref{sec:mt}, we further include Mamba-2-based versions in Figure~\ref{fig:mamba2-mt-alti-vs-avg}.

\subsection{Retrieval-based Generation}
\label{subsec:app-rag}

\paragraph{Passkey Retrieval.} We compute accuracy statistics in the passkey retrieval task for Mamba-2 370M for each variation (1, 2 and 4 passkeys). We observe that the model has a heavy bias towards the first passkey that appears in context as the average accuracy decreases as more keys get introduced (Table~\ref{tab:passkey_acc}). To strengthen our argument, accuracy heavily depends on whether the desired passkey is the first that appears (Table~\ref{tab:passkey_firstkey}).

\begin{table}[t]
\centering
\small
\setlength{\tabcolsep}{4pt} %
\begin{tabular}{lcccccc}
\toprule
\multirow{2}{*}{\textbf{Size}} & \multicolumn{3}{c}{\textbf{1024}} & \multicolumn{3}{c}{\textbf{2048}} \\
\cmidrule(lr){2-4} \cmidrule(lr){5-7}
& \textbf{$k=1$} & \textbf{$k=2$} & \textbf{$k=4$} & \textbf{$k=1$} & \textbf{$k=2$} & \textbf{$k=4$} \\
\midrule
130M & 99.8 & 58.2 & 28.1 & 99.7 & 57.3 & 30.8 \\
370M & 100.0 & 57.6 & 33.1  & 98.0 & 55.1 & 34.1 \\
780M & 99.8 & 68.1 & 60.2 & 84.5 & 59.3 & 51.4 \\
1.4B & 99.3 & 63.2 & 39.6 & 99.7 & 60.8 & 38.9 \\
\bottomrule
\end{tabular}
\caption{Mamba-2 accuracy (\%) in the Passkey Retrieval task at recovering the correct output. We vary the model size, sequence length (1024 and 2048) and the number of keys $k \in \{1, 2, 4\}$. Computed over 1000 samples. }
\label{tab:passkey_acc}
\end{table}

\begin{table}[t]
\centering
\small
\setlength{\tabcolsep}{4.5pt} %
\begin{tabular}{lcccccc}
\toprule
\multirow{2}{*}{\textbf{Size}} & \multicolumn{2}{c}{\textbf{2 Passkeys}} & \multicolumn{2}{c}{\textbf{4 Passkeys}}  \\
\cmidrule(lr){2-3} \cmidrule(lr){4-5}
& \textbf{First} & \textbf{Second} & \textbf{First} & \textbf{Second+} \\
\midrule
130M & 74.3 & 41.2  & 46.9 & 22.2  \\
370M & 65.4 & 47.3  & 53.6 & 26.7  \\
780M & 76.9 & 59.1  & 82.0 & 53.4  \\
1.4B & 81.7 & 43.6  & 64.4 & 31.8  \\
\bottomrule
\end{tabular}
\caption{Mamba-2 accuracy (\%) in the Passkey Retrieval task at recovering the correct key if the correct key is the \textit{First} to appear or the \textit{Second+} to appear. Computed over 1000 samples of length 1024.}
\label{tab:passkey_firstkey}
\end{table}

\begin{figure*}[t]
  \centering 
  \includegraphics[width=0.495\textwidth]{figs/rag/niah-1k-2keys-att-word-17-cropped.pdf}
  \includegraphics[width=0.495\textwidth]{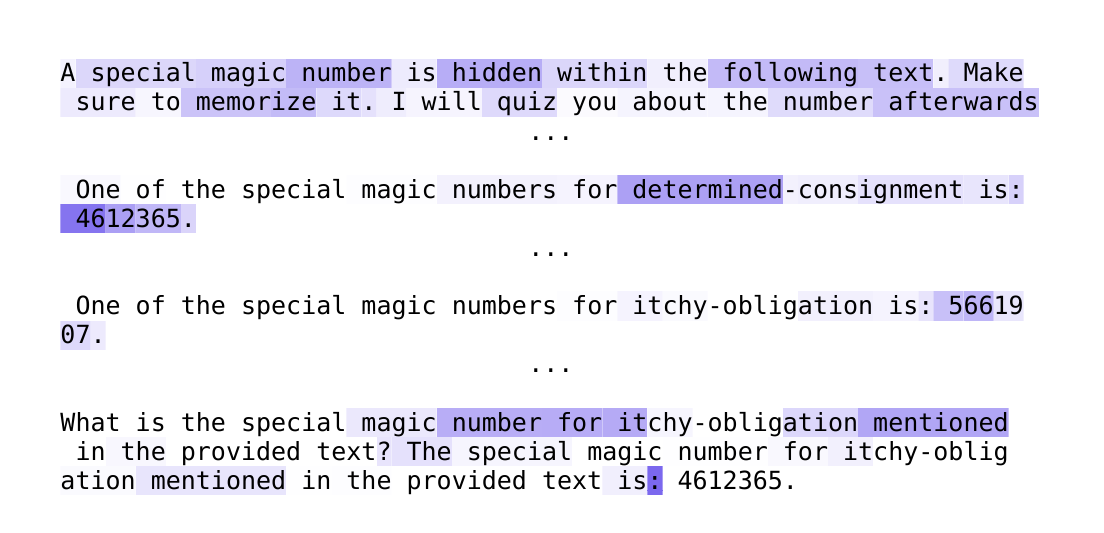}
  \caption{Attention plots obtained by \methodname ($\ell_2$) (left) and MambaAttention (right) on a Passkey Retrieval sample, showing that MambaAttention focuses on several misleading tokens, such as ``determined'', ``number for'', and ``mentioned''. In contrast, \methodname ($\ell_2$) focuses only on meaningful strings, like ``4612365'' (the predicted key) and ``5661907'' (the correct key).}
  \label{fig:app-mamba-passkey}
\end{figure*}

\begin{figure*}[t]
  \centering
  \includegraphics[width=0.495\textwidth]{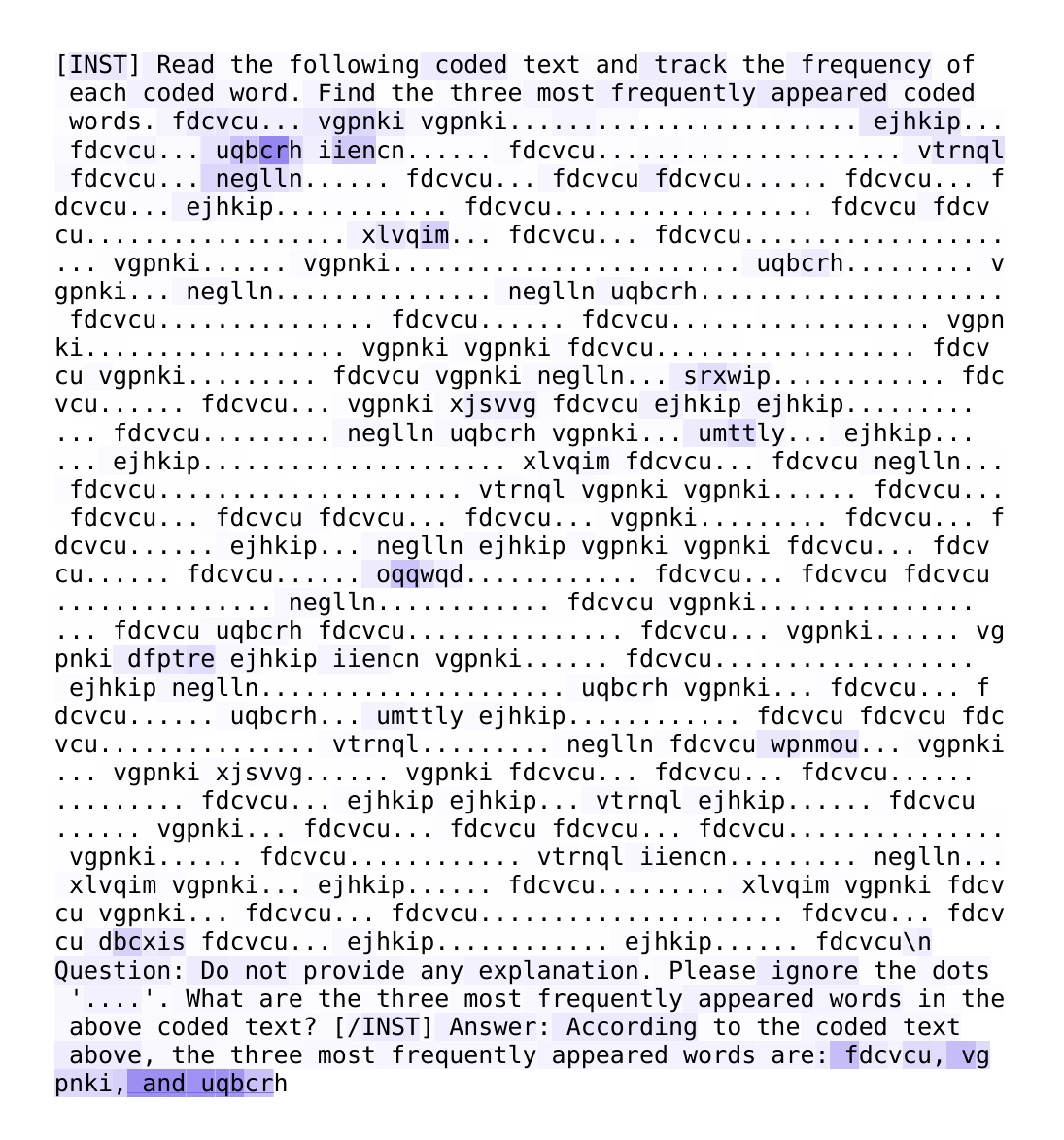}
  \includegraphics[width=0.495\textwidth]{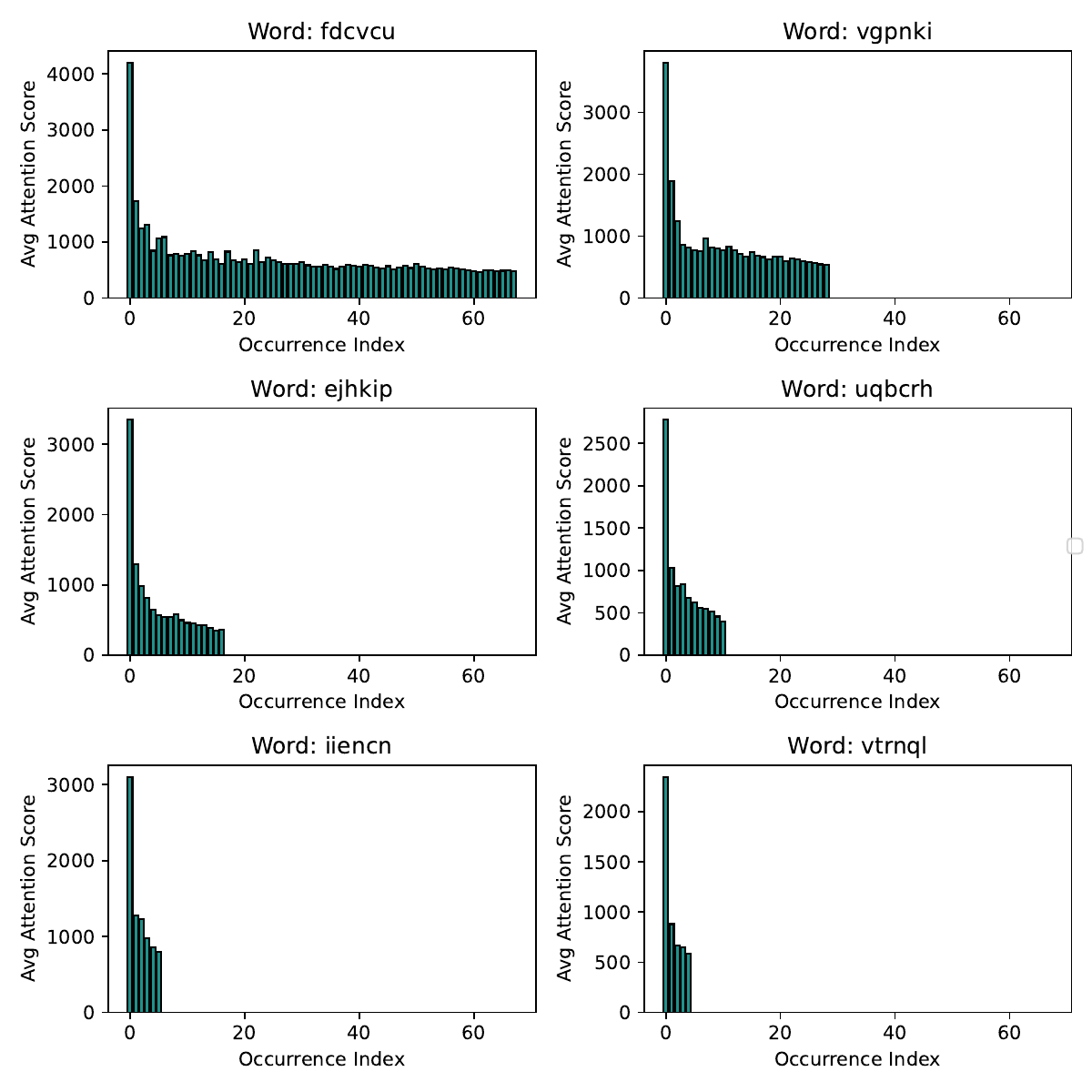}
    \label{fig:mamba-fwe-occurrences}
  \caption{Left: Attention map obtained by \methodname ($\ell_2$) on the Frequent Word Extraction task, showing that the model is focusing on the incorrectly generated ``and uqbcr'' token range (Mamba-2 370M layer 23). 
  Right: Average attention score per word instance, showing that the model's focus reduces heavily after the first few word occurrences.}
  \label{fig:mamba-fwe}
\end{figure*}

\paragraph{Frequent Word Extraction.} In Figure~\ref{fig:mamba-fwe} (left) we plot Mamba-2's focus over the context tokens in the Frequent Word Extraction task when we only consider the ``and uqbc'' (underlined) tokens. As we can see, only some words get attended to, making it difficult for the model to track word frequencies. To strengthen this effect, the average attention per word instance decreases heavily. For example, the word ``fdcvcu'' occurs 68 times and its first few occurrences have an average attention score across layers substantially higher than the remainder. 

\section{AI Assistants}
We used Cursor during development, and ChatGPT during paper writing for grammar correction.

\end{document}